\documentclass[11pt]{article}
\usepackage[utf8]{inputenc}
\usepackage{graphicx}
\usepackage{booktabs}
\usepackage{pdflscape}
\usepackage{adjustbox}
\usepackage{multicol}

\usepackage{amsmath,amssymb}
\usepackage{microtype} 
\usepackage[T1]{fontenc}
\usepackage{hyperref}
\usepackage{authblk}
\usepackage{ragged2e}
\usepackage{titlesec}
\usepackage{multirow}
\usepackage[none]{hyphenat}  
\usepackage{natbib} 
\usepackage{setspace} 
\usepackage{blindtext}
\usepackage[a4paper, total={7in, 10in}]{geometry}
\usepackage{hyperref} 
\usepackage[most]{tcolorbox}

\newtcolorbox{highlightblock}{
  breakable,              
  enhanced,
  colback=blue!40,       
  colframe=blue!40,      
  boxrule=0pt,             
  sharp corners,
  interior style={opacity=0.6}, 
    }

\hypersetup{
    colorlinks = true,      
    linkcolor  = blue,      
    citecolor  = magenta,   
    urlcolor   = red       
}

\title{A Systematic Review of Key Retrieval-Augmented Generation (RAG) Systems:\\
Progress, Gaps, and Future Directions\\[2em]}
\usepackage{authblk}

\author[1]{Agada Joseph Oche\thanks{Corresponding author: \texttt{joe88data1@gmail.com}}}
\author[1]{Ademola Glory Folashade\thanks{\texttt{gloryademola112@gmail.com}}}
\author[2]{Tirthankar Ghosal}
\author[3]{Arpan  Biswas}

\affil[1]{Bredesen Center for Interdisciplinary Research, University of Tennessee, Knoxville, USA, 37996}
\affil[2]{National Center for Computational Sciences, Oak Ridge National Laboratory, Oak Ridge, USA, 37830}
\affil[3]{University of Tennessee-Oak Ridge Innovation Institute, University of Tennessee, Knoxville, USA, 37996}

\date{\today\\[3em]}

\begin{document}

\maketitle

\begin{abstract}
Retrieval-Augmented Generation (RAG) represents a major advancement in natural language processing (NLP), combining large language models (LLMs) with information retrieval systems to enhance factual grounding, accuracy, and contextual relevance. This paper presents a comprehensive systematic review of RAG, tracing its evolution from early developments in open-domain question answering to recent state-of-the-art implementations across diverse applications. The review begins by outlining the motivations behind RAG, particularly its ability to mitigate hallucinations and outdated knowledge in parametric models. Core technical components—retrieval mechanisms, sequence-to-sequence generation models, and fusion strategies—are examined in detail. A year-by-year analysis highlights key milestones and research trends, providing insight into RAG’s rapid growth. The paper further explores the deployment of RAG in enterprise systems, addressing practical challenges related to retrieval of proprietary data, security, and scalability. A comparative evaluation of RAG implementations is conducted, benchmarking performance on retrieval accuracy, generation fluency, latency, and computational efficiency. Persistent challenges such as retrieval quality, privacy concerns, and integration overhead are critically assessed. Finally, the review highlights emerging solutions, including hybrid retrieval approaches, privacy-preserving techniques, optimized fusion strategies, and agentic RAG architectures. These innovations point toward a future of more reliable, efficient, and context-aware knowledge-intensive NLP systems.
\end{abstract}

\noindent\textbf{Keywords:} Retrieval Augmented Generation (RAG), Large Language Model (LLM), Generative AI, Natural Language Model (NLP)

\newpage

\begin{multicols}{2}
\justifying    
\sloppy        

\section{Introduction}
Since its formal introduction in the seminal work of \citep{Lewis2020} in 2020, Retrieval-Augmented Generation (RAG) has witnessed rapid advancements, marked by a significant surge in research interest and scholarly publications. This paper offers a unique and comprehensive review of the key developments and contributions in the field to date. The remainder of this introduction outlines the background and motivation for this review, defines its scope and objectives, and provides an overview of the paper's organization.

\subsection{Background and Motivation}
Large-scale pre-trained language models have demonstrated an ability to store vast amounts of factual knowledge in their parameters, but they struggle with accessing up-to-date information and providing verifiable sources. This limitation has motivated techniques that augment generative models with information retrieval. \textit{Retrieval-Augmented Generation} (RAG) emerged as a solution to this problem, combining a neural retriever with a sequence-to-sequence generator to ground outputs in external documents \citep{Lewis2020}. The seminal work of \citep{Lewis2020} introduced RAG for knowledge-intensive tasks, showing that a generative model (built on a BART encoder–decoder) could retrieve relevant Wikipedia passages and incorporate them into its responses, thereby achieving state-of-the-art performance on open-domain question answering. RAG is built upon prior efforts in which retrieval was used to enhance question answering and language modeling \citep{Lee2019, Guu2020, Karpukhin2020}. Unlike earlier extractive approaches, RAG produces free-form answers while still leveraging non-parametric memory, offering the best of both worlds: improved factual accuracy and the ability to cite sources. This capability is especially important to mitigate \textit{hallucinations} (i.e., believable but incorrect outputs) and to allow knowledge updates without retraining the model \citep{Lewis2020, IBM2023}. 

Since its introduction, RAG has gained significant attention in both research and industry. A growing body of literature has extended RAG with improved retrievers and generators, and the approach has been applied to a wide range of domains. By 2023, the RAG paradigm underpinned hundreds of research publications and numerous commercial systems \citep{Gao2024RAGSurvey, Merritt2023}. In academia, researchers have scaled up retrieval-augmented models and refined their architectures—examples include leveraging larger pre-trained models with retrieval in the loop \citep[e.g.,][]{Izacard2022, Borgeaud2022}. In parallel, industry adoption of RAG has been swift: leading tech companies have integrated retrieval-augmented generators into search engines, virtual assistants, and enterprise question-answering applications \citep{IBM2023, Merritt2023}. RAG now powers applications from open-domain QA and customer support chatbots to tools that automatically generate answers with supporting evidence. This broad adoption underscores the significance of RAG as a foundation for making generative AI more reliable and knowledge-aware. This paper provides a unique perspective on to review of literature in RAG by providing detailed yearly research progress in RAG, developing new perspectives, and evaluating trends.

\subsection{Scope and Objectives}
The objective of this systematic review is to provide a comprehensive overview of the development of RAG and its expanding role in information access. We aim to answer several key research questions: (1) \textit{How has RAG been progressed every year since its inception, and what are the major technical milestones in its research and deployment?} (2) \textit{What challenges and solutions have emerged for integrating RAG with proprietary or private data sources, and what gaps remain (e.g., in security and privacy)?} (3) \textit{How has RAG been used in accelerating material discovery and characterization} (4) \textit{How are RAG systems categorized and how does this categorization affect their performance?}. By addressing these questions, the review seeks to chart the evolution of RAG, evaluate its current capabilities and limitations, and identify areas for future work.

Over the past few years, progress in RAG has been marked by continuous innovation and new applications. We chronicle the advancement year-by-year, highlighting important academic contributions and industry developments that have shaped the field. Special attention is given to the integration of RAG with \textit{proprietary data}—an area of growing interest as organizations apply RAG to internal knowledge bases. This involves examining techniques for efficient retrieval on private corpora and the handling of sensitive information, as well as open issues around data privacy \citep{Zeng2024}. Recent systems have also demonstrated that users can interact with an RAG-powered agent to obtain information directly from the web or a document corpus, rather than through traditional ranked search results \citep{Nakano2022, Shuster2022}. This paradigm blurs the line between search engine and dialogue agent, opening questions about usability, accuracy, and trust in such interfaces. Overall, the review considers this and also consolidates knowledge on how RAG techniques have matured and what objectives remain for future research and development.

\subsection{Paper Organization}
The remainder of this paper is structured as follows. \textbf{Section~2 (Methodology)} explains the review methodology, including the literature search strategy, inclusion/exclusion criteria, and approach to data synthesis. \textbf{Section~3 (Foundations of RAG)} provides a technical overview of retrieval-augmented generation, describing its core components (retrievers, indexes, generators) and the baseline architectures introduced by seminal works. \textbf{Section~4 (Year-by-Year Progress)} presents a chronological synthesis of RAG developments from 2017 onward, highlighting key research milestones.  \textbf{Section~5 (RAG for Proprietary data and Industry Implementation)} examines enterprise implementation of RAG on proprietary data by key industry players. \textbf{Section~6 (RAG Systems Evaluation)} benchmarks different RAG implementations and variants, summarizing their performance across standard datasets and tasks. \textbf{Section~7 (Challenges of RAG Systems)}, \textbf{Section~8 (Discussion and Future Direction)} and \textbf{Section~9 (Conclusion)} finally provide current research gaps and potential future directions to expand the applications to various domain problems.

\section{Methodology}
\label{sec:methodology_no_math}

This section details the systematic review methodology employed to survey RAG papers. It comprises three main steps:
(1)~designing a \textbf{Search Strategy} to capture a wide range of relevant works, (2)~defining \textbf{Inclusion and Exclusion Criteria} to refine the initial corpus, and (3)~implementing a \textbf{Data Extraction and Synthesis} process to analyze and consolidate findings.

\subsection{Search Strategy}
To ensure comprehensive coverage, we searched both academic and
industry-focused literature on RAG. Multiple digital libraries were queried, including ACL Anthology, IEEE Xplore, ACM Digital Library, and Google Scholar. We included documents published from 2017 up to the end of mid 2025,
covering early “retrieve-and-generate” approaches and more recent RAG-specific techniques.

\paragraph{Keywords and Databases.}
We used a set of pre-defined keywords, such as “\emph{retrieval-augmented generation (RAG)},” “\emph{dense retrieval},” “\emph{hybrid retrieval LLM},” “\emph{RAG proprietary data},” and “\emph{LLM web search}.” These
queries captured works ranging from open-domain QA to secure enterprise implementations. Each keyword search was executed on the above-listed databases, resulting in a pool of references that included journal articles, conference papers, technical reports, and white papers.

\paragraph{Initial Screening.}
A comprehensive list of potentially relevant works was formed by merging all search results and removing duplicates. Abstracts and titles were checked to confirm alignment with the RAG focus. If a work concentrated solely on retrieval or generation in isolation, without discussing how these components integrate, it was set aside for possible exclusion.

\subsection{Inclusion and Exclusion Criteria}
We next applied a formal screening process to determine which references genuinely contributed insights into RAG. The criteria below guided our decisions:

\subsubsection{Inclusion Criteria}
\begin{itemize}\setlength{\itemsep}{0pt}
     \item\textbf{Relevant to RAG or closely related baselines}: Works that clearly integrated a retriever with a generative language model or used retrieval to supply context to a generator \citep{Lewis2020}.

\item\textbf{Knowledge-Intensive Tasks}: Studies centered on open-domain QA, fact-checking, knowledge-grounded dialogue, or other tasks benefiting from external document retrieval. 

 \item\textbf{Peer-Reviewed or Reputable Sources}: Publications presented at major AI/NLP venues (ACL, NeurIPS, ICML, EMNLP) or recognized industrial R\&D labs (e.g., IBM, Meta, NVIDIA)
\citep{IBM2023, Merritt2023}.

\item\textbf{Preprint}: Preprints are also included to broaden the scope of the survey.

\item\textbf{English Language}: For consistency and to support thorough evaluation, only English texts were included.
end
\end{itemize}

\subsubsection{Exclusion Criteria}
\begin{itemize}\setlength{\itemsep}{0pt}
     \item\textbf{Solely Retrieval or Solely Generation}: Articles focusing strictly on IR techniques or purely generative models without explicit retrieval-augmented integration were not included.
\item\textbf{Minimal Discussion of RAG}: Any mention of
retrieval+generation was peripheral or superficial, lacking
substantial results or analyses.
\item\textbf{Non-Substantive Publications}: Very short abstracts, publicity notes, or materials without verifiable methodology were excluded.
\item\textbf{Non-English Papers}: Not considered due to feasibility constraints.
\end{itemize}

Based on these criteria, the initial corpus was refined into a
finalized set of documents deemed pertinent to the state-of-the-art in RAG.

\subsection{Data Extraction}
\label{subsec:dataextraction_no_math}
For each included publication, we collected key information, such as basic bibliographic details, the retrieval method (e.g., dense vs. sparse), the generator architecture (e.g., T5, BART, GPT), and the evaluated tasks or datasets. This allowed us to systematically compare different RAG implementations and their reported performance. We also looked for and extracted information on the challenges facing RAG implementation. The survey is not limited to peer-reviewed journal articles and conference proceedings, preprints, technical reports, and industry white papers were also reviewed. The review covers the application of RAG systems in all domains 
\paragraph{Synthesis Process.}
All extracted details/data were gathered in a central repository, allowing cross-study comparisons. We grouped research outputs by year of publication to track the chronological evolution of RAG, highlighting seminal breakthroughs and subsequent expansions. In line with systematic review principles, we combined both qualitative (themes, research directions) and quantitative (performance figures, latency measures) observations.
\paragraph{Ensuring Reliability.}
Disagreements during the review were resolved through discussion or by consulting a third reviewer. This final step ensured consistent application of the inclusion/exclusion criteria and reliable data extraction. The data collected then served as the foundation for our analysis in subsequent sections, including discussions on year-by-year progress, enterprise
applications,and proposed solutions.

\section{Foundations of RAG}
\subsection{Definition and Key Concepts}
\paragraph{Retrieval-Augmented Generation (RAG)}: RAG is a framework that combines a neural text \textbf{retrieval} module with a text \textbf{generation} module to improve the quality of generated responses in knowledge-intensive tasks. Formally, a RAG model augments a sequence-to-sequence (seq2seq) generator with access to an external text corpus (non-parametric memory) via a retriever \citep{Lewis2020, Karpukhin2020}. Given an input query $x$, the retriever $R$ selects a small subset of relevant documents $Z = \{z_1, z_2, \dots, z_K\}$ from a large corpus $\mathcal{C}$ (with $K \ll |\mathcal{C}|$) \citep{Karpukhin2020}. The generator then conditions on both the query $x$ and the retrieved documents $Z$ to produce an output $y$ (such as an answer or a descriptive text). Formally, the RAG model can be viewed as a latent variable generative model that defines a probability distribution over outputs $y$ by marginalizing over the retrieved documents $z_i$:
\begin{equation}\label{eq:rag-marginal}
    P(y \mid x) \;=\; \sum_{i=1}^{K} P_{\text{ret}}(z_i \mid x)\; P_{\text{gen}}(y \mid x, z_i)\,,
\end{equation}
where $P_{\text{ret}}(z_i \mid x)$ is the probability of retrieving document $z_i$ given query $x$ (the retriever's output distribution), and $P_{\text{gen}}(y \mid x, z_i)$ is the generator's conditional probability of producing $y$ given $x$ and a particular retrieved document $z_i$. In practice, $P_{\text{ret}}(z_i \mid x)$ is typically non-zero only for the top-$K$ retrieved items, providing a tractable approximation to the full sum over the corpus \citep{Lewis2020}. The retriever $R$ itself can be defined as a function $R(x, \mathcal{C}) \to Z$ that takes a query and returns a small subset $Z$ of corpus $\mathcal{C}$ (with $|Z|=K \ll |\mathcal{C}|$) likely to contain information relevant to $x$ \citep{Karpukhin2020}. By design, RAG models maintain two kinds of \textbf{memory}: a \emph{parametric memory} (the knowledge encoded in the generator’s weights) and a \emph{non-parametric memory} (the external text corpus accessed via retrieval) \citep{Lewis2020}. A standard RAG architecture is illustrated in Figure~\ref{fig:rag_pipeline} below. 
A key distinction between RAG and pure large language model (LLM) generation is the use of this external non-parametric knowledge source at inference time. Traditional LLM-based generation relies solely on the model’s internal parameters for knowledge, which can lead to \textbf{hallucinations} and factual inaccuracies when the model’s training data does not adequately cover the query’s topic \citep{Lewis2020}. In contrast, RAG explicitly grounds the generation of retrieved documents that serve as up-to-date evidence, enabling the model to generate content supported by those documents. This retrieval step means that RAG’s outputs can be more accurate and \textbf{factually correct} compared to generation from a standalone LLM, especially for knowledge-intensive queries. Empirically, \citep{Lewis2020} demonstrates that a RAG model generates more specific and factual responses than a parametric-only generator, since the retrieved text provides verified information that the generator can incorporate. Another benefit is that the knowledge in a RAG system can be easily \textbf{updated} by modifying the document index (or corpus) without retraining the generator, addressing the stiffness of LLMs that have fixed knowledge up to their training cutoff date. In summary, RAG introduces a modular architecture where a retrieval component supplies relevant context “just in time” for the generator, marrying the strengths of Information Retrieval (IR) with those of large-scale generation.
\end{multicols}
\begin{figure}[h]
    \centering
    \includegraphics[width=0.85\textwidth]{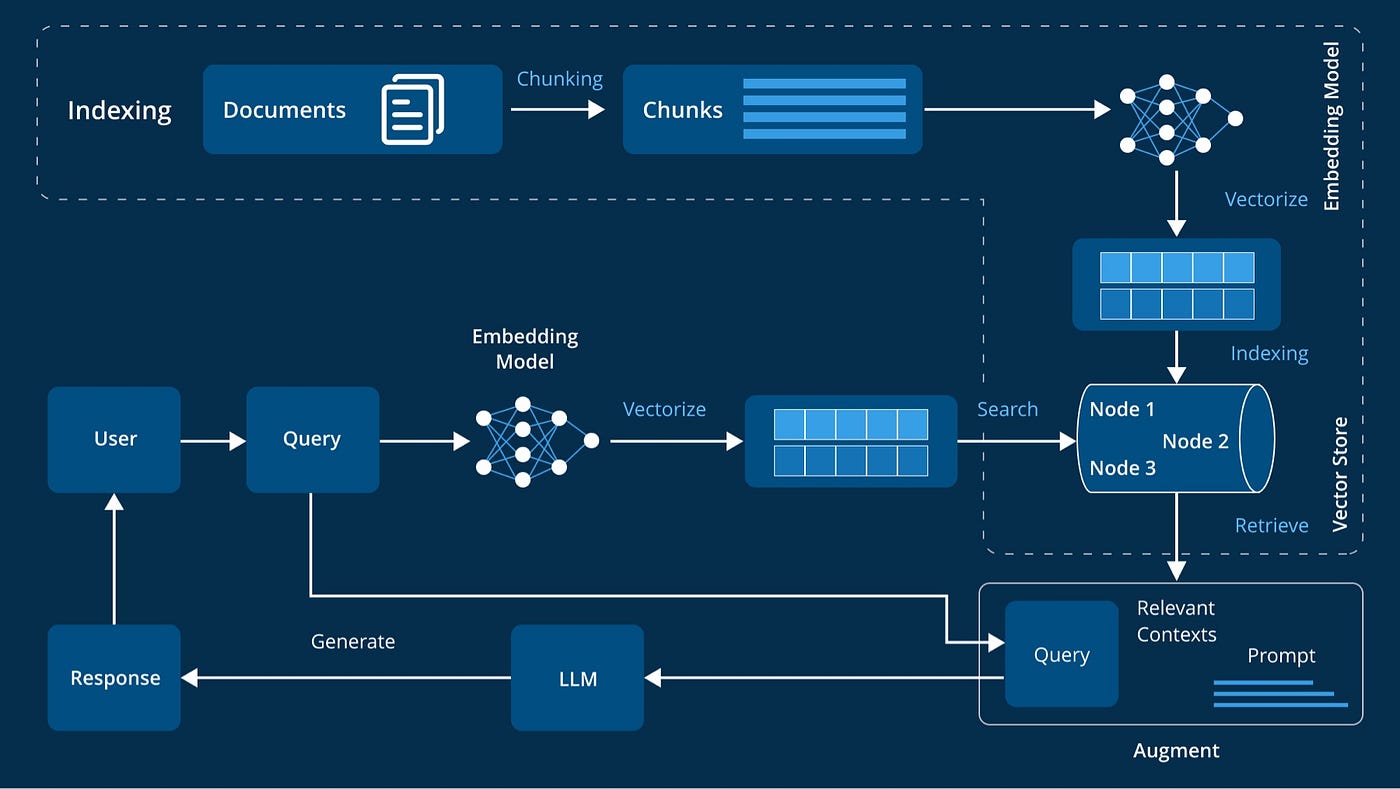}
    \caption{Illustration of a RAG Architecture.}
    \label{fig:rag_pipeline}
\end{figure}

\begin{multicols}{2}

\paragraph{Chunking, Embedding, and  (Re)ranking}:A typical RAG pipeline consists of four stages: chunking, embedding, (re)ranking, and generation. First, chunking is applied to the knowledge source: large documents are segmented into smaller, self-contained pieces (e.g., paragraphs or passages) for indexing. Using fine-grained text chunks as retrieval units improves the chance that a query will surface a highly relevant fragment, rather than an entire lengthy document \citep{Lewis2020}. For example, open-domain QA systems split Wikipedia articles into passage chunks to enable pinpoint retrieval of answer-containing segments \citep{Karpukhin2020, Lewis2020}. The chunk size is typically tuned to balance context completeness and specificity – chunks must be large enough to contain useful context, yet small enough to match queries narrowly and fit within model context windows. Next, each chunk is embedded into a high-dimensional vector representation that encodes its semantic content. This is usually done with a transformer-based bi-encoder that produces dense vector embeddings of text \citep{Karpukhin2020}. The embeddings serve as keys in a vector index (or vector database) that supports efficient nearest-neighbor search. At query time, the user’s query is likewise embedded into the same vector space, and the system performs similarity search to retrieve the most relevant chunk vectors. In contrast to sparse keyword search, dense embeddings enable semantic matching: a query about “financial earnings” can retrieve a chunk about “quarterly revenue” even if exact words differ \citep{Karpukhin2020}. Modern RAG implementations often combine dense retrieval with lightweight filtering or hybrid search (e.g., BM25 + embeddings) to improve recall for difficult queries. The result of the initial retrieval stage is a candidate set of top-$k$ chunks that are potentially relevant to the query. To further improve precision, an optional re-ranking step is applied on the retrieved candidates before generation. The top-$k$ chunks from the first stage may contain some irrelevant or only tangentially related items, since embedding similarity is a coarse proxy for relevance. A re-ranker model (typically a cross-encoder transformer that jointly encodes query and document) evaluates each retrieved chunk in the context of the query and produces a refined relevance score \citep{Nogueira2019}. By re-scoring and sorting the candidates, the re-ranker ensures that the most pertinent chunks (for example, those actually containing the answer to a question) are ranked highest. This two-stage retrieval process – a fast dense retriever followed by a more accurate but expensive re-ranker – has been shown to significantly boost retrieval performance on knowledge-intensive benchmarks. For instance, neural cross-attention re-rankers achieve substantially higher accuracy than single-stage retrievers alone \citep{Nogueira2019}. In practice, re-ranking is crucial in high-stakes applications (e.g., legal or medical QA), where one must maximize the likelihood that the top context passages truly address the user’s query. After re-ranking, the top $N$ (e.g. 3–5) chunks are selected as the final context passages for the generative model. In the generation stage, the LLM produces an answer or response conditioned on the retrieved external chunks. Typically, a sequence-to-sequence model (such as T5 or BART) is used so that the retrieved text can be prepended or incorporated into the model’s input along with the user query \citep{Lewis2020, Raffel2020}. During training, the model learns to copy or attend to the relevant facts from the retrieved documents and integrate them into a coherent output. This approach allows the generator to cite up-to-date, specific information beyond its parametric knowledge. For example, \citep{Lewis2020} show that a RAG model (BART-based) can accurately answer open-domain questions by retrieving and conditioning on Wikipedia text, dramatically reducing hallucinations compared to a standalone LLM. The generated output can also include references to source documents, providing traceability for the facts used. Retrieval augmentation thus serves as a “live memory” for the LLM: it supplies factual grounding from an external knowledge base while the language model creates fluent and contextually relevant text. Notably, recent large-scale studies have demonstrated that even very large models benefit from retrieval augmentation. For instance, the RETRO model augments a 7.5-billion-parameter transformer with a database of trillions of tokens, yielding improved perplexity and factual accuracy by looking up passages during generation \citep{Borgeaud2022}. In summary, chunking, embedding, re-ranking, and generation work in concert in RAG systems to leverage external knowledge – the retrieval components identify and prioritize relevant information, and the generation component uses that information to produce answers that are both informative and grounded in source data. This modular design has become a foundation for building more reliable and explainable AI assistants in knowledge-intensive domains.

\subsection{Technical Components of RAG}
A RAG system is composed of two primary components – a \textbf{retriever module} and a \textbf{generator module} – along with a strategy for fusing their outputs. We break down these components and the underlying mechanics as follows.

\paragraph{Retriever Module (Dense Passage Retrieval).} The retriever’s job is to efficiently identify which pieces of text in a large corpus are relevant to the input query $x$. Modern RAG implementations typically use \textbf{dense retrievers} like Dense Passage Retrieval (DPR) \citep{Karpukhin2020} in lieu of traditional keyword search. In DPR, a bi-encoder architecture is employed: a \textit{question encoder} $E_q(x)$ maps the query $x$ to a $d$-dimensional vector, and a \textit{passage encoder} $E_p(d)$ maps each candidate document (or passage) $d$ in the corpus to a $d$-dimensional vector in the same space. Relevance is assessed via a similarity score, usually the dot product of these vectors: 
\[
\text{sim}(x, d) \;=\; E_q(x)^\top\, E_p(d)\,. 
\]
At query time, the retriever computes $v_q = E_q(x)$ and then finds the top-$K$ documents whose vector $E_p(d)$ has highest inner product with $v_q$. This search for maximum inner product can be implemented efficiently using Approximate Nearest Neighbor techniques (e.g., FAISS) to handle millions of documents in sub-linear time. The output of the retriever is the set $Z = \{z_1, \ldots, z_K\}$ of top-ranked documents and potentially their similarity scores. One can view the retriever as defining a probability distribution $P_{\text{ret}}(z \mid x)$ over documents $z$ in the corpus, such that:
\begin{equation}\label{eq:retriever-softmax}
    P_{\text{ret}}(z \mid x) \;\propto\; \exp\big( E_q(x)^\top E_p(z)\big)\,,
\end{equation}
with the normalization $\sum_{d \in \mathcal{C}} \exp(E_q(x)^\top E_p(d))$ (in practice approximated by summing over retrieved candidates rather than all of $\mathcal{C}$). In other words, the retriever assigns higher probability (or score) to documents whose embedding is most similar to the query’s embedding. The DPR retriever is usually first trained on pairs of questions and relevant passages to ensure $E_q$ and $E_p$ produce representations that maximize dot-products for true Q–A pairs and minimize them for irrelevant pairs. The training objective is often a \textbf{contrastive loss}: for a given question $q$ with a gold relevant passage $p^+$ and a set of negative passages $\{p^-_1, \dots, p^-_N\}$, the encoder is trained to maximize $\text{sim}(q,p^+)$ while minimizing $\text{sim}(q,p^-)$ for negatives. This can be formulated as a cross-entropy loss treating the positive passage as the correct class among one positive and $N$ negatives:
\begin{equation}
\resizebox{0.95\hsize}{!}{$
    \mathcal{L}_{\text{ret}}(q, p^+) = -\log 
    \frac{\exp(\text{sim}(q, p^+))}{\exp(\text{sim}(q, p^+)) + 
    \sum_{j=1}^{N} \exp(\text{sim}(q, p^-_j))}
$}
\end{equation}

which encourages $E_q$ and $E_p$ to embed true pairs closer together than any negative pair \citep{Karpukhin2020}. After training, the retriever can generalize to new queries: it embeds the query and efficiently finds the nearest neighbor passages in the index. This retrieval step is crucial because it narrows down the evidence from potentially billions of tokens to a manageable subset that the generator will actually consider.

\paragraph{Generator Module (Conditional Seq2Seq Model).} The generator in a RAG pipeline is typically a sequence-to-sequence language model that produces the final answer or output text, given the input query and retrieved documents. Formally, the generator defines a conditional distribution $P_{\text{gen}}(y \mid x, Z)$ over output sequences, where $Z = \{z_1,\ldots,z_K\}$ are the retrieved passages. The generator is often initialized from a pre-trained transformer-based seq2seq model (such as BART or T5) to leverage rich language generation capabilities \citep{Lewis2020}. During generation, the model is provided with the question (or prompt) as well as the content of the retrieved documents. There are multiple ways to feed the retrieved context to the generator:
\begin{itemize}\setlength{\itemsep}{0pt}
    \item In an \textbf{early fusion} approach, one can concatenate the query $x$ with the text of all retrieved passages into a single extended input sequence (perhaps with special separators) and have the seq2seq model attend to all of it at once. This is straightforward: the generator effectively treats the combined text as context and learns to pick out the relevant bits when producing the answer. However, this can be difficult if $K$ is large, as the input may become very long.
    \item In the \textbf{late fusion} approach adopted by RAG \citep{Lewis2020}, the generator considers one retrieved document at a time as a context and then marginalizes over the document choices (as in Eq.~\ref{eq:rag-marginal}). Specifically, the RAG-Sequence variant fixes a single document $z_i$ as context for generating the entire output and computes $P(y \mid x)$ by summing the probabilities $P(y \mid x,z_i)$ weighted by the retriever’s confidence $P(z_i \mid x)$. Another variant, RAG-Token, allows the generator to switch between documents at the token level, effectively marginalizing over the document choice for each generated token \citep{Lewis2020}. In both cases, the generator $P_{\text{gen}}(y \mid x, z)$ itself works like a standard seq2seq model: it factorizes over the output tokens $y_1, \dots, y_T$ as $\prod_{t=1}^{T} P_\theta(y_t \mid x, z, y_{<t})$, i.e. it generates one token at a time, attending to the input query $x$ and the content of $z$ (or multiple $z$’s if early fusion). The generator’s architecture typically uses an encoder-decoder Transformer: the encoder encodes the combination of $x$ and $z$, and the decoder produces $y$ autoregressively.
\end{itemize}
Because the generator is conditioning on retrieved evidence, it tends to produce outputs that are supported by that evidence. For example, if the query asks for a specific factual answer, the generator can copy or rephrase the needed information from one of the retrieved passages. This is in contrast to a vanilla language model which would attempt to rely on parametric knowledge (which might be outdated or incomplete). In sum, the generator module in RAG is responsible for fluent and coherent text generation, but crucially it is \emph{grounded} by the retrieval context, which guides it toward accurate content.

\paragraph{Fusion Mechanisms and Answer Aggregation.} A critical aspect of RAG systems is how to fuse information from multiple retrieved documents $z_1, \dots, z_K$ when producing the final answer. Different fusion strategies have been explored:
- \textbf{Marginalization (Probabilistic Fusion):} As described, RAG treats the retrieved documents as latent variables and marginalizes over them \citep{Lewis2020}. This means the model doesn’t commit to one retrieved source up front; instead, it considers each in turn and combines their contributions by summing probabilities. Concretely, if $y$ is an output sequence, a RAG model might compute its probability by $P(y|x) = \sum_{i=1}^K P_{\text{ret}}(z_i|x)\,P_{\text{gen}}(y|x,z_i)$ (Eq.~\ref{eq:rag-marginal}). During training, this encourages the model to distribute probability mass across any document that could yield the correct answer, reinforcing multiple evidence paths. At inference, one can approximately marginalize by taking the most likely $y$ under this mixture model.
- \textbf{Direct concatenation (Early Fusion):} As mentioned, another approach is to simply feed all top-$K$ retrieved texts into the generator at once (often referred to as “Fusion-in-Decoder” when implemented in a decoder-attention context). In this setup, the generator effectively performs its own internal fusion by attending over a combined context. This approach has the advantage that the generator can directly cross-attend to multiple documents and integrate their content, but it may require a more powerful model to handle very long concatenated inputs. It also does not explicitly model the per-document probabilities $P(z_i|x)$.
- {Weighted Aggregation:} Some systems introduce an attention or weighting mechanism over retrieved documents. For instance, the generator’s decoder might assign different attention weights to different passages at each decoding step, effectively learning which source is most useful for generating the next token. This can be seen as a soft fusion: rather than hard marginalization or simple concatenation, the model dynamically blends information. In practice, approaches like \citep{Lewis2020} found that marginalization (which is a form of weighting by the retriever’s scores) works well, especially when the retriever is accurate. Other works have since experimented with learnable fusion weights or iterative retrieval-generation cycles, but the core idea is the same: the model must reconcile possibly conflicting or complementary information from multiple documents to produce a single, coherent answer.

The choice of fusion affects the system’s ability to handle conflicting evidence and the credit assignment during training (i.e., which document gets “credit” for a correct answer). RAG’s probabilistic fusion provides a principled way to train the retriever and generator together by marginalizing, whereas direct concatenation treats the problem in a single forward pass of a generator (often fine for tasks where evidence is mostly additive or when using very large generators). Fusion strategies continue to be an active area of research, but they all serve the goal of effectively utilizing multiple retrieved pieces of text to improve answer completeness and correctness.

\paragraph{Training and Optimization.} Training a RAG model involves objectives for both the retriever and the generator, which can be combined in an end-to-end manner. A common training approach is as follows: first, pre-train or initialize the retriever on a relevance task and initialize the generator on a language modeling or seq2seq task (often using a pre-trained model checkpoint). Then, perform joint fine-tuning on the target task (e.g., a QA dataset or a knowledge-intensive dialogue dataset) by maximizing the likelihood of the correct output $y^*$ given the input $x$ and allowing gradients to flow into both the generator and retriever. The training objective for the whole RAG system can be written as the expected negative log-likelihood:
\[
\mathcal{L}_{\text{RAG}} = -\log P(y^* \mid x)\,,
\] 
where $P(y^* \mid x)$ is computed as in Eq.~\ref{eq:rag-marginal}. Because $P(y^*|x)$ is a sum over documents, the gradient will encourage whichever retrieved documents $z_i$ that helped predict $y^*$ (by giving high $P_{\text{gen}}(y^*|x,z_i)$) to have their retrieval probability $P_{\text{ret}}(z_i|x)$ increased. In effect, the model learns to adjust the retriever to fetch better supporting documents and adjust the generator to rely on them appropriately. This joint training is typically done with standard backpropagation; since the retriever’s selection operation is not differentiable for all documents, one uses the top-$K$ approximation (only those contribute to the loss) and treats the retrieval probabilities for those as soft variables. \citep{Lewis2020} report that initializing the retriever with DPR and then fine-tuning end-to-end yields the best results, as opposed to training from scratch. Notably, the retriever is trained indirectly here: it does not receive explicit labels of which document is correct, but the generator’s success or failure on producing $y^*$ provides a supervision signal. This is sometimes called “self-supervised” retriever training or “feedback” training.

In addition to end-to-end training, various optimization tricks may be used: e.g., using a small learning rate for the retriever if it’s already strong, or alternating between retriever-focused and generator-focused updates. In some cases, researchers have also explored \textbf{contrastive learning at the generation level} (to reduce ambiguity between retrieved passages) or reward-based objectives if the task is not a straightforward next-word prediction. However, the most common training objective for RAG is the simple maximum likelihood training of the seq2seq model, augmented by the latent document marginalization. The result is a system where both components are tuned to the end task: the retriever learns to bring useful evidence, and the generator learns to incorporate that evidence into the output. This joint optimization is a major advantage of RAG over non-integrated pipelines, as it aligns the retriever’s objective with generating correct final answers (not just retrieving vaguely related documents).

\subsection{Historical Context}
The evolution of RAG builds upon earlier developments in open-domain question answering (QA) and neural information retrieval. Traditional open-domain QA systems were typically pipeline-based, consisting of a retrieval step followed by a reading or extraction step \citep{Chen2017WikipediaQA}. For example, \citep{Chen2017WikipediaQA} introduced the DrQA system, which first used a TF-IDF or BM25 retriever to select Wikipedia articles and then fed those to a machine reader model to extract answers. This established the value of retrieving relevant text from a large corpus as an essential first step in answering open-domain questions. However, in such pipeline approaches the retriever was not integrated into the learning of the reader, and the system could not adjust retrieval based on the end task’s needs. Subsequent research sought to bridge this gap by jointly learning retrieval and answering. Notably, the concept of using learned dense representations for retrieval emerged as a powerful alternative to traditional sparse retrieval. Early milestones in this direction include \textit{latent retrieval} models: \citep{Lee2019} proposed the ORQA model, which treats retrieval as a latent variable problem and pre-trains a neural retriever on an unsupervised “inverse cloze task” before jointly fine-tuning it with a reader on QA. Around the same time, \citep{Guu2020} introduced REALM, a retrieval-augmented language model pre-training method that incorporated a differentiable retriever into the pre-training of a masked language model. REALM demonstrated that pre-training a model to retrieve and reason over Wikipedia could significantly improve open-domain QA, highlighting the benefit of coupling a language model with a learned retrieval mechanism \citep{Guu2020}. These efforts were focused primarily on question answering (often extractive), but they laid important groundwork for retrieval-augmented generation by showing that retrieval and neural text generation can be trained in tandem.

In parallel, the idea of combining external knowledge with neural networks has roots in earlier \textbf{memory-augmented models}. For instance, Memory Networks \citep{Weston2015} and subsequent variants allowed neural networks to read from an external memory of facts and use that information to answer questions or generate responses. These models (e.g., \citep{Weston2015, Sukhbaatar2015}) demonstrated the feasibility of non-parametric memory for reasoning, albeit on smaller-scale knowledge bases or synthetic tasks. While memory network architectures were often task-specific and required the memory to be relatively small or structured, they presaged the RAG approach by emphasizing that not all knowledge needs to be baked into model parameters—some can be looked up as needed. Another line of work in dialogue systems also integrated retrieval into generation: the \textbf{Wizard of Wikipedia} project \citep{Dinan2019} is a prime example, where a conversational agent retrieves relevant Wikipedia sentences and conditions a generative dialogue model on those sentences to produce knowledgeable responses. This retrieval-based dialogue system (published in 2019) demonstrated improved factuality and depth in conversational responses, reflecting the general trend that augmenting generators with retrieved context yields more informative and correct outputs.

These developments converged in 2020 with the formalization of Retrieval-Augmented Generation by \citep{Lewis2020}, who unified the retriever-reader architecture with seq2seq generation in an end-to-end framework. The RAG model of \citep{Lewis2020} was a culmination of insights from open-domain QA and neural IR: it used a \textbf{dense passage retriever} \citep{Karpukhin2020} to fetch text chunks from Wikipedia and a powerful seq2seq generator (BART) to produce answers or summaries, training both components jointly. By marginalizing over multiple retrieved documents (as in Eq.~\ref{eq:rag-marginal}), the RAG system could leverage several pieces of evidence and was shown to outperform both parametric-only models and earlier retrieve-and-read pipelines on knowledge-intensive tasks like open QA \citep{Lewis2020}. The introduction of RAG in 2020 is considered a key milestone because it generalized retrieval-augmented architectures beyond QA to any generative task requiring external knowledge. It also spurred a new line of research into \textbf{knowledge-enhanced text generation}, influencing subsequent models that further refined retrieval modules, document ranking, and fusion techniques for even better performance.

\paragraph{Execution Flow of a RAG System.} To summarize the interactions of these components, we outline the step-by-step execution flow of a typical RAG system processing a query:
\begin{enumerate}
    \item \textbf{Query Encoding:} Given an input query $x$ (e.g., a question in natural language), the retriever’s query encoder $E_q$ first encodes $x$ into a vector representation $v_q$. This vector captures the semantic meaning of the query in a dense embedding space.
    \item \textbf{Document Retrieval:} Using the query vector $v_q$, the system performs a search over the document index (the external corpus $\mathcal{C}$). It computes similarity scores $\text{sim}(x, d)$ for documents $d$ (often via inner product with pre-computed document embeddings $E_p(d)$) and retrieves the top $K$ documents with highest scores. These top-$K$ documents $Z = \{z_1, ..., z_K\}$ are assumed to be the most relevant pieces of text related to the query.
    \item \textbf{Context Preparation:} The text of the retrieved documents $Z$ is retrieved from the knowledge store. The RAG system now has access to these $K$ passages (e.g., paragraphs from Wikipedia) that can serve as supporting context. Depending on the fusion strategy, the system either concatenates these passages or will handle them separately in the next step.
    \item \textbf{Answer Generation:} The query $x$ along with the retrieved context $Z$ are fed into the generator model. If using early fusion, all of $Z$ (or a subset, if some cutoff like $K'$ is used) is given as additional input to the encoder. If using late fusion (as in the original RAG), the generator will consider each $z_i$ in $Z$ in turn. The generator’s decoder then produces an output sequence $y$ (the answer or response). During this decoding, the model may attend to relevant parts of the retrieved texts. For example, the decoder might focus on a specific retrieved passage that contains a needed fact when generating the corresponding part of the answer. In RAG’s late fusion, the decoder actually generates an answer for each $z_i$ implicitly and the probabilities of tokens are combined by marginalization. In practice, one can sample or beam-search for the best output $y$ across the combined evidence. 
    \item \textbf{Fusion and Output:} If multiple candidate outputs were considered (e.g., one per retrieved document), the model marginalizes or otherwise aggregates them to produce the final answer. Often the single most likely sequence $y$ is selected as the output. The final answer $y$ is then returned by the system as the response to the query. Optionally, the system might also output the passages it used (providing provenance or justification, which is a useful feature of RAG systems).
\end{enumerate}
This flow involves interleaving retrieval and generation in a seamless way. Notably, steps 1–2 (retrieval) drastically reduce the problem space by focusing on a handful of documents out of potentially millions, and steps 3–5 ensure that the information in those documents is synthesized into a fluent answer. The entire process is typically very fast at inference: encoding the query and searching the index can be done in tens of milliseconds with efficient vector databases, and generation is on the order of the length of the output (with modern transformers generating dozens of tokens per second). Therefore, RAG systems can scale to handle user queries in real-time applications, all while maintaining higher accuracy by leveraging updated and explicit knowledge. The end-to-end design means that if the output is incorrect, the system can be improved by either enhancing the retriever (fetch more relevant docs) or the generator (better use of docs), or both, which aligns with the modular evaluation and training typical in IR+NLP pipeline but now integrated into a single model.

\section{Year-by-Year Progress in RAG}
\label{sec:progress}
RAG, as it is known today, was proposed in \citep{Lewis2020}, but before then, the retrieval and read pipeline, which operates like RAG. This section reviews the evolution of RAG from the pre RAG erra till date. The anual year-by-year progress of RAG is illustrated in Figure~\ref{fig:rag_evolution}. 

\end{multicols}
\begin{figure}[h]
    \centering
    \includegraphics[width=0.85\textwidth]{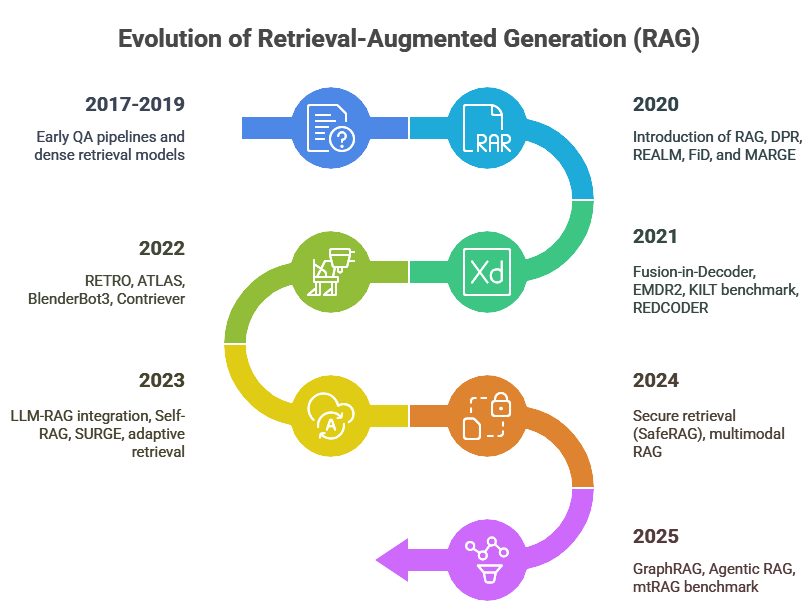}
    \caption{Evolution of a RAG Architecture.}
    \label{fig:rag_evolution}
\end{figure}

\begin{multicols}{2}

\subsection{Initial Proposals and Early Research (2017--2019)}
Before the term \emph{retrieval-augmented generation (RAG)} was coined, researchers explored methods to combine information retrieval with neural models for question answering (QA) and text generation. Early open-domain QA systems typically employed a \emph{retrieve-and-read} pipeline: a search module locates relevant documents, then a neural
reader model extracts or generates answers \citep{Chen2017WikipediaQA}. For instance, the 2017 \emph{DrQA} framework answered questions using Wikipedia as a knowledge source by pairing a TF-IDF-based document retriever with an RNN-based reader trained to extract answer spans. Although \citep{Chen2017WikipediaQA} demonstrated strong performance on open-domain trivia tasks, these systems were piecemeal in nature: the retrieval and generation components were not trained jointly, and the
end-to-end approach was limited to extractive answers. In 2018, work shifted toward tighter integration between retrieval and reading. \citep{wang2018r3} proposed \(\mathrm{R^{3}}\) (\emph{Reinforced Reader-Ranker}), adding a neural ranker to score retrieved passages by answer likelihood. The system then learned ranker--reader synergy via reinforcement learning, boosting open-domain QA accuracy by better filtering relevant evidence. Meanwhile, neural IR
methods gained traction. \citep{Lee2019} introduced
\emph{Latent Retrieval} in their \emph{Open-Retrieval QA} (ORQA) framework, training a \emph{dense} retriever and a reader end-to-end with only question--answer supervision. Specifically, the retriever embeddings were pretrained on an inverse cloze task and then adapted to select evidence documents that help the QA model answer correctly. This
concept of \emph{dense retrieval}, which outperformed sparse BM25 by up to 19\% in exact-match QA scores, would become the foundation of subsequent RAG models. Still, these early systems were limited mostly to extractive QA, without a unified end-to-end training for generative
outputs.
\subsection{Major Milestones (2020--2024)}
\paragraph{2020 --- Birth of RAG.} The year 2020 marked a turning point with the official formalization of
\emph{retrieval-augmented generation}. \citep{Lewis2020} coined the term ``RAG'' and demonstrated its power on knowledge-intensive tasks. RAG explicitly splits knowledge across (i)~a neural \emph{retriever} and (ii)~a neural \emph{generator}, each playing a distinct role. Given a query \(x\), RAG retrieves top-\(k\) relevant passages
\(\{z_1,\dots,z_k\}\) from a large text corpus (via a learned dense index) and then conditions a seq2seq model on both \(x\) and
\(\{z_i\}\). Mathematically,
\begin{equation*}
  P(y \mid x) \;=\; \sum_{i=1}^{k} P_{\theta}(y \,\mid\, x, z_i)\;\,P_{\phi}(z_i \,\mid\, x),
\end{equation*}
where \(P_{\phi}(z_i \mid x)\) is the retriever's distribution over
documents (often realized via a softmax on 

\begin{figure}[h]
    \centering
    \includegraphics[width=1\textwidth]{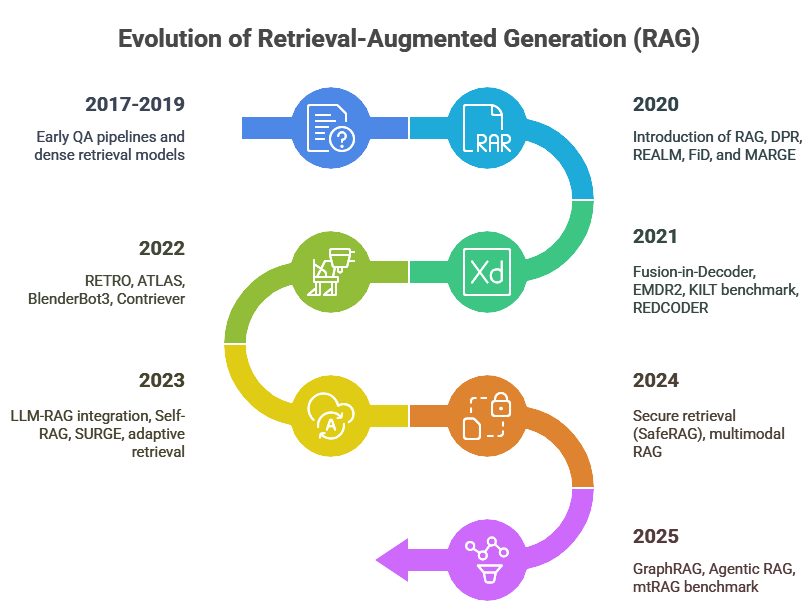}
    \caption{Evolution of RAG: 2017 - Mid-2025.}
    \label{fig:rag_evolution}
\end{figure}

inner product scores), and
\(P_{\theta}(y \mid x,z_i)\) is the conditional probability of the
generator. \citep{Lewis2020} used a BART-based generator and a Dense
Passage Retriever \citep{Karpukhin2020}, outperforming older retrieve-and-read
pipelines by leveraging both parametric and non-parametric memory. Simultaneously, \citep{Guu2020} introduced \emph{REALM}:
retrieval-augmented \emph{pretraining}, which integrated a differentiable
retriever into a language model to predict masked tokens with retrieved evidence. REALM achieved significant QA gains over conventional LMs, validating that external knowledge injection helps both at pretraining and fine-tuning. \citep{Karpukhin2020} also released their Dense
Passage Retrieval (DPR) approach, establishing a simple but effective dual-encoder architecture. By 2020's end, retrieval-augmented strategies had become key to state-of-the-art QA, and generative models demonstrated improved factuality by referencing retrieved text. Some particular progress made in 2020 are discussed below:

\paragraph{Dense neural retrieval.}  Classical TF–IDF/BM25 retrieval limited earlier pipeline QA systems.  \citep{karpukhin2020dpr} introduced \emph{Dense Passage Retrieval (DPR)}, training dual BERT encoders to embed questions and passages into a shared space.  DPR improved top 20 recall by 9–19 pp over BM25 and became the de facto index for RAG models.

\paragraph{End‑to‑end retriever–generator architectures.}  Building on DPR, \citep{lewis2020retrieval} proposed the eponymous \emph{RAG} model: a BART generator conditions on $k$ passages retrieved (and jointly trained) via DPR.  Two variants—RAG‑Sequence and RAG‑Token—achieved state‑of‑the‑art exact‑match scores on Natural Questions and TriviaQA, while producing more specific, better‑grounded answers than closed‑book T5 of comparable size.  Concurrently, the \emph{Fusion‑in‑Decoder (FiD)} architecture of \citep{Izacard2021} showed that a T5 decoder attending over dozens of retrieved passages could push QA accuracy even higher, underscoring that modern seq2seq models can synthesise evidence from many documents.

\paragraph{Retrieval‑aware pre‑training.}  Rather than bolt retrieval on during fine‑tuning, REALM \citep{guu2020realm} trained a BERT‑style model to retrieve Wikipedia passages to fill masked tokens, optimising retriever and LM jointly.  REALM outperformed larger closed‑book models by up to 16 pp in open QA.  MARGE \citep{lewis2020marge} extended the idea: the model reconstructs a target document from related texts it retrieves, yielding strong zero‑shot results in multilingual summarisation and translation.

\paragraph{Applications beyond QA.}  RAG principles generalized quickly.  In knowledge‑grounded dialogue, bridging a prior (inference‑time) and posterior (training‑time) retriever improved response relevance \citep{chen2020dialogue}.  Fact‑checking systems retrieved evidence then generated verdicts, while early experiments hinted at factuality gains in abstractive summarisation.  The \emph{KILT} benchmark \citep{petroni2021kilt} unified eleven knowledge‑intensive tasks with a shared Wikipedia snapshot and evaluation that scores both answer correctness and evidence retrieval.  A single RAG baseline proved competitive across QA, dialogue, fact verification, and slot‑filling, highlighting RAG’s domain‑agnostic promise.

\paragraph{Impact and open questions.}  
By year‑end 2020, RAG systems of only a few hundred million parameters were surpassing 11‑billion‑parameter closed‑book LMs \citep{Lewis2020,roberts2020}, demonstrating the efficiency of hybrid parametric + non‑parametric memory.  Challenges that remained—in retrieval latency at scale, multi‑hop reasoning, and tighter faithfulness evaluation—set the agenda for subsequent work.  Nonetheless, the 2020 breakthroughs firmly established retrieval‑augmented generation as a core methodology for building knowledgeable, transparent, and updatable NLP systems.

\paragraph{2021 --- Improving Generation Quality.}
Work in 2021 centered on refining how retrieval and generation interact, leading to \emph{Fusion-in-Decoder} (\emph{FiD}) by
\citep{Izacard2021}. Rather than marginalize over top-\(k\) passages as in RAG, FiD concatenates them and feeds them all into a T5-based seq2seq, allowing the decoder to attend to multiple documents simultaneously. This architecture demonstrated that large generative models can effectively combine evidence from many passages, achieving further gains on open QA benchmarks. Meanwhile, new tasks beyond QA
surfaced, such as fact-checking, knowledge-grounded dialogue, and entity-rich tasks \citep{petroni2021kilt}, all relying on retrieval to supply correct and up-to-date information. By 2021, RAG had expanded its reach to a wider range of knowledge-intensive NLP domains. Some major progress made in 2021 are discussed below:

\paragraph{}

Open-domain QA was an early focus of RAG research. \cite{Izacard2021} introduced the Fusion-in-Decoder (FiD) architecture, which encodes each retrieved passage independently and fuses them in the decoder. FiD achieved state-of-the-art results on benchmarks such as Natural Questions and TriviaQA, showing that performance improves as more evidence passages are. In parallel, \citep{sachan-etal-2021-end} developed an end-to-end training scheme (EMDR2) for multi-document QA. By treating retrieval as a latent variable and iteratively training the retriever and reader with an EM-like algorithm, they improved answer accuracy across datasets, establishing new state-of-the-art results without supervised retrieval labelsyielded. Together, these works demonstrated that advanced RAG architectures and training strategies yield substantial QA gains in 2021.

\paragraph{}

RAG techniques also permeated knowledge-grounded dialogue system. The KILT benchmark introduced by \citep{petroni-etal-2021-kilt} unified several knowledge-intensive tasks (QA, fact-checking, dialogue, etc.) on a common Wikipedia knowledge source. KILT showed that a shared dense retriever and generative model can serve as a strong generalist system: it outperformed task-specific baselines in fact-checking and open-domain QA, and was competitive on dialogue and entity linking:contentReference[:4]{index=4}. This suggests that RAG-style models (a retriever plus a seq2seq generator) can effectively support multi-turn, knowledge-grounded conversation. While Wizard-of-Wikipedia (2019) had earlier demonstrated retrieval in dialogue, the KILT results in 2021 underscored that unified RAG pipelines are robust across dialogue and QA domains.

\paragraph{}

In summarization and content generation, RAG has been applied to integrate relevant context. One example is code summarization: \citep{parvez-etal-2021-retrieval-augmented} proposed REDCODER, a framework that retrieves relevant code snippets or summaries to augment a code generation model. By searching a codebase for similar examples, REDCODER improved both code generation and summarization quality on Java and Python benchmarks. More broadly, retrieval-augmented summarization was noted to help produce more accurate and up-to-date summaries by grounding language models in external documents (e.g.\ news or scientific articles), though in 2021 most demonstrations were in specialized domains like code or clinical summarization. Fact-checking similarly benefited: retrieving evidence from text (e.g.\ news or scientific literature) provides the basis for verifying model outputs. KILT explicitly includes fact-checking (the FEVER dataset) in its suite, and found that a general RAG approach is competitive on evidence retrieval for claims.

\paragraph{}

Several methodological improvements emerged in 2021. Beyond FiD and EMDR2, researchers explored how to structure RAG models. \citep{lewis2020retrieval} distinguished two RAG variants: one conditions on a fixed set of retrieved passages for the entire output, while the other can fetch different passages at each decoding step. \citep{Izacard2021} exemplified the former by concatenating encoded passages in the decoder. but in 2021 the focus was on multi-passage integration. Additionally, enhancements to retrievers were important: dense retrievers like DPR \citep{karpukhin2020dpr} underpin many 2021 RAG systems, often pretrained on open-domain QA. Training retrievers jointly with generators, as in \citep{sachan2021emdr2}, improved retrieval relevance. Some studies also integrated topic or dialogue context into retrieval for conversational settings. In all cases, the synergy of retrieval and generation architectures was a key theme.

\paragraph{}

By the end of 2021, standardized benchmarks and code releases helped consolidate RAG progress. The KILT benchmark in particular brought together 11 tasks across multiple knowledge-intensive domains. On this benchmark, the combination of a pretrained retriever and a large generative model achieved strong performance across QA, dialogue, and even slot filling, highlighting the versatility of RAG pipelines. In open QA, the Natural Questions, TriviaQA, and EfficientQA datasets continued to serve as metrics for RAG-based methods; FiD and other models raised the bar on these tasks in 2021. Summaries of RAG results noted major gains over prior baselines in both accuracy and factuality.

\paragraph{2022 --- Scaling \& Specialization.}
In 2022, \citep{Borgeaud2022} introduced \emph{RETRO}
(\emph{Retrieval-Enhanced Transformer}), revealing that a moderately sized 7.5B-parameter model could match GPT-3 (175B params) if it retrieves relevant text chunks from a huge corpus (2T tokens), thereby proving that retrieval can \emph{substitute for scale}. RETRO thus illustrated the efficiency advantage of external knowledge over
solely increasing model parameters. In parallel, \citep{Izacard2022} presented \emph{Atlas}, a retrieval-augmented language model aiming for \emph{few-shot learning} on knowledge-intensive tasks. Atlas combined a
T5-based generator with an advanced dense retriever, pre-trained on massive unlabeled text. Demonstrating strong results in few-shot scenarios (only 64 examples needed for respectable performance), Atlas highlighted the synergy between retrieval and pre-trained seq2seq in
reducing reliance on large labeled datasets.

\citep{Lewis2020} and DPR \citep{karpukhin2020dpr} also proved that conditioning generation on retrieved passages boosts factual accuracy. DeepMind’s \textbf{RETRO} model \citep{borgeaud2022retro} attaches a nearest‑neighbour retriever to every decoding window of a 7‑billion‑parameter transformer; with a 2‑trillion‑token index, RETRO matches GPT‑3 performance while using 25$\times$ fewer parameters, underscoring retrieval’s efficiency gains.

\paragraph{}

Google’s \textbf{ATLAS} \citep{Izacard2022} unifies retrieval and generation during \emph{pre‑training}.  Using the unsupervised \textbf{Contriever} dense retriever \citep{izacard2022contriever} and a Fusion‑in‑Decoder reader, ATLAS achieves new state‑of‑the‑art results on Natural Questions and TriviaQA and, in few‑shot mode, outperforms much larger PaLM‑540B by 3 EM on NQ with only 64 examples.  These results highlight that high‑quality retrieval plus multi‑document fusion can outperform sheer parameter count.

\paragraph{}

Beyond QA, RAG became core to knowledge‑grounded \emph{dialogue}.  Meta’s \textbf{BlenderBot 3} \citep{shuster2022bb3} couples a 175B LLaMA‑style generator with live internet search and long‑term memory, reducing hallucinations and increasing user engagement in open‑domain conversation.  BlenderBot 3’s deployment study shows that users prefer retrieval‑grounded responses and that continual online learning can keep the retriever index current. Retrieval quality itself improved through unsupervised contrastive training.  Contriever \citep{izacard2022contriever} dispenses with labelled (question, passage) pairs, yet surpasses BM25 and even DPR on BEIR benchmarks, making high‑recall indexes available for any corpus.  Such retrievers power ATLAS and other 2022 RAG systems, demonstrating that scalable training data is no longer a bottleneck.

\paragraph{}

Finally, 2022 work extended RAG to \emph{fact‑checking, summarization, and few‑shot learning}.  ATLAS gains 5F1 on FEVER, indicating that retrieved evidence helps verdict generation.  Early studies in retrieval‑augmented summarization show improved factual consistency by grounding summaries in external documents.  Few‑shot evaluations reveal that retrieval narrows the data gap: ATLAS and RETRO deliver strong accuracy with under 100 task examples, whereas closed‑book baselines require orders of magnitude more data.

\paragraph{Outlook}  
By the end of 2022, RAG had broadened from open‑domain QA into a general recipe for knowledge‑intensive NLP.  Parameter‑efficient hybrids like RETRO and ATLAS challenge the notion that bigger models alone yield better knowledge; instead, high‑quality retrieval and multi‑document reasoning emerge as key levers.  Open challenges include faster retrieval over trillion‑token corpora, differentiable multi‑hop reasoning, and robust evaluation of evidential faithfulness, but 2022 firmly established retrieval‑augmented generation as a premier path toward up‑to‑date, factual, and data‑efficient language models.

\paragraph{2023 --- RAG Meets LLMs.}
By 2023, mainstream LLM-based applications (e.g., ChatGPT with plugins, Bing Chat, enterprise chatbots) widely incorporated retrieval \citep{hadi2023survey}. This \emph{retrieve-then-generate} paradigm was used to mitigate
\emph{hallucinations} and update factual knowledge post-training. The debate emerged as to whether \emph{long-context LLMs} (with tens of thousands of tokens) could negate the need for retrieval systems. Studies like
\citep{li2024rag_vs_long} showed that while large context windows can absorb more text, RAG remains more \emph{cost-efficient} and better at exposing citations. Hybrid approaches also arose: letting a model choose
between retrieving or just using a long context. Overall, RAG became a cornerstone for credible LLM deployments needing up-to-date knowledge and interpretability. SOme of the areas that received attention in 2023 are discussed below:

\textbf{Scale and few-shot learning:}
Atlas\citep{izacard2023atlas}, an 11B-parameter retrieval-augmented model, achieved 42.4\% accuracy on Natural Questions with only 64 training examples, outperforming a 540B closed-book model by 3\%. Atlas also set new few-shot records on TriviaQA and FEVER (gains of +3–5\%), matching 540B-scale performance on multi-task benchmarks. Crucially, Atlas’s dense document index can be easily updated with new text, demonstrating updatable knowledge.

\textbf{Adaptive retrieval:} 
Self-RAG\citep{asai2023selfrag} trains a single language model to generate special “reflection” tokens that trigger on-demand retrieval and self-critique. In experiments, 7B and 13B Self-RAG models substantially outperformed ChatGPT and a RAG-augmented Llama-2-chat baseline on open-domain QA, reasoning, and fact-verification tasks, yielding much higher factual accuracy and citation precision.

\textbf{Knowledge-grounded dialogue:} 
Retrieval augments dialogue systems to improve consistency and informativeness. Kumari et al.\citep{kumari2023dialog} incorporate retrieved persona and context snippets in long conversation modeling, showing that adding relevant knowledge improves response quality. Similarly, Kang et al.\citep{kang2023surge} propose \emph{SURGE}, which retrieves relevant subgraphs from a knowledge graph and uses them to bias the response generation. SURGE produces more coherent, factual responses grounded in the retrieved knowledge.

\textbf{Summarization and explanation:} 
RAG has been applied to summarization and explanation tasks. By retrieving source documents or evidence passages, RAG-augmented summarizers produce more accurate and detailed summaries than closed-book models. Likewise, in fact-checking pipelines, retrieving evidence before verification leads to more reliable verdicts and explanations. These applications extend RAG’s grounding advantages beyond QA to a broader range of generative tasks.

\textbf{Benchmarks and evaluation:}
\\Knowledge-intensive benchmarks track RAG progress. OpenQA tasks (Natural Questions, TriviaQA, HotpotQA) and the KILT benchmark suite (including QA, fact-checking, slot filling, etc.) are standard evaluation sets. In 2023, RAG models dominated many KILT tasks and few-shot QA challenges. New evaluation tools also emerged: for example, the RAGAS framework provides reference-free metrics for RAG pipelines, and the RAGTruth corpus (Niu et al.\ 2024) enables fine-grained analysis of hallucinations in RAG outputs. 2023 also witnessed a major spike in publications and adoption of RAG. Some other works are on Active RAG \citep{jiang-etal-2023-active}, Improving domain adaptation of RAG models for open question answering \citep{siriwardhana2023improving}, content filtering for RAG \citep{wang2023learning}, and RAG with self-memory 
\citep{cheng2023lift}. 

\paragraph{2024 --- Recent Advances.} 
In 2024, research on RAG continues to push on \emph{secure retrieval}
frameworks, multi-hop reasoning, and domain specialization. Several groups explore \emph{differentiable retrievers} that can be tuned in an end-to-end pipeline, while others
investigate merging large-context attention with retrieval indexing. Meanwhile, new techniques aim to reduce the chance the generator ignores retrieved data or merges contradictory documents incorrectly. RAG-based chat systems in healthcare, finance, and law now incorporate advanced fact-checking modules to ensure that only \emph{vetted} external sources
influence the final output. Some notable works in 2024 are Evaluating Retrieval Quality in RAG \citep{salemi2024evaluating}, Benchmarking RAG for Medicine \citep{xiong2024benchmarking}, Benchmarking LLM in RAG \citep{chen2024benchmarking}, Review of RAG for AI Generated Content \citep{zhao2024retrieval}, Unifying Context Ranking with RAG \citep{yu2024rankrag}, Searching for the Best Practice in RAG \citep{wang2024searching}, Finetuning Vs RAG for Less Popular Knowledge \citep{soudani2024fine}, Integrating RAG wit LLM in Nephrology \citep{Miao2024NephrologyRAG}, RAG for Copyright Protection \citep {golatkar2024cpr}, RAG for Textual Graph Understanding and Question Answering \citep{he2024g}, Interactive AI with RAG for NExt Generation Networking \citep{zhang2024interactive}, Web Application for RAG: Implementation and Testing \citep{radeva2024web}, Overcoming Challenges for Long Input in RAG \citep{jin2024long}, and Adapting Language Model for Domain Specific RAG \citep{zhang2024raft}. Some specifics in RAG development as at 2024 are discussed below:

\textbf{Infrastructure and standardized evaluation:}
The community recognised a need for common tooling and shared tasks. \emph{Ragnarök} introduced a reusable end‑to‑end framework and provided industrial baselines for the inaugural \textbf{TREC 2024 RAG Track} \citep{pradeep2024ragnarok}. Beyond code, evaluation methodology itself became a research focus: \textbf{ARAGOG} proposed automatic grading of RAG outputs that correlates with human judgements, analysing retrieval precision and answer similarity across advanced pipelines \citep{eibich2024arag}. These efforts mark a shift from anecdotal demos to systematic, reproducible assessment.

\textbf{Adaptive retrieval and self‑reflection:}
Building on ideas such as Self‑RAG, several 2024 works taught models to \emph{decide when—and how much—to retrieve}. SAM‑RAG dynamically filters documents and verifies both evidence and final answers in multimodal contexts, improving accuracy without unnecessary retrieval calls \citep{zhai2024samrag}. For complex visual‑question‑answering, \emph{OmniSearch} plans multi‑hop retrieval chains on the fly, demonstrating large gains on the new Dyn‑VQA benchmark \citep{li2024omnsearch}. These results confirm that retrieval policies, not just retriever quality, matter for difficult queries.

\textbf{Multimodal RAG breaks out:}
Where earlier RAG research was text‑only, 2024 saw a surge in multimodal extensions. SAM‑RAG and OmniSearch both combine text and image evidence, while concurrent frameworks (e.g.\ mR$^{2}$AG and M3DocRAG) introduce retrieval–reflection loops or structured vision–language indexes. Surveys published this year chart the design space and highlight open issues such as cross‑modal alignment and vision‑aware reranking \citep{zhai2024samrag, li2024omnsearch}.

\textbf{Progress in dense retrieval:}
High‑recall retrievers remain the backbone of every RAG system. 2024 research emphasised \emph{unsupervised or instruction‑tuned} retrievers that avoid costly labelled data, building on contrastive pre‑training techniques and LLM‑augmented embedding models.  
These retrievers power the top submissions in the TREC track and underpin production deployments discussed in industrial white papers.

\textbf{Rapid growth and forward outlook:}
A bibliometric snapshot counted more than 1,200 RAG‑related papers on arXiv in 2024 alone \citep{zhao2024retrieval}, compared with fewer than 100 the previous year, underscoring the field’s rapid maturation.  
Looking ahead, challenges include ultra‑fast retrieval over trillion‑token corpora, faithfulness verification for multi‑hop reasoning, and energy‑efficient multimodal indexing.  
Nevertheless, 2024 firmly established RAG as the default strategy for grounding large language (and vision‑language) models in up‑to‑date, attributable knowledge.

\subsection{2025 --- The Current Direction}
The community is exploring how to marry graph knowledge with text retrieval.  A comprehensive survey formalised the \emph{GraphRAG} paradigm and mapped design choices for graph‑aware retrievers and generators \citep{han2025graphrag}.  
A companion study compared vanilla RAG and GraphRAG across QA and summarisation, showing complementary strengths and proposing hybrid fusion strategies \citep{han2025ragvgraph}.  

\textbf{Security and robustness.}  
\emph{SafeRAG} introduced the first security benchmark for RAG pipelines, cataloguing four attack classes (silver noise, context conflict, soft‑ad, DoS) and demonstrating that 14 representative systems fail even simple manipulations \citep{liang2025saferag}.  
Findings sparked interest in provenance tracking and adversarially trained retrievers.

\textbf{Agentic and selective retrieval.}  
A survey on \emph{Agentic RAG} synthesised emerging patterns—reflection, planning, tool use—and argued that autonomous agents can orchestrate multi‑hop retrieval more effectively than static pipelines \citep{singh2025agentic}.  
Concrete instantiations followed:  
\begin{itemize}
  \item \emph{SIM‑RAG} learns a self‑skeptic critic that decides when to stop multi‑round search, cutting redundant calls and boosting exact‑match on five QA sets by up to 4pp\citep{yang2025simrag}.  
  \item \emph{Self‑Routing RAG} jointly trains an LLM to choose between internal knowledge and external retrieval, reducing retrieval volume 29\% while raising accuracy 5pp\citep{wu2025srrag}.  
\end{itemize}

\textbf{Conversational evaluation at scale.}  
IBM’s \emph{mtRAG} benchmark is filling a gap in multi‑turn assessment: 110 human‑written conversations (7.7 turns each) across four domains, plus synthetic variants, revealed that state‑of‑the‑art systems struggle with unanswerable and follow‑up questions \ citep {katsis2025mtrag}.  
Alongside automatic judges such as RL‑F and RB‑LLM, mtRAG enables holistic measurement of retrieval, generation, and turn‑level faithfulness.

\textbf{Emerging directions.}  
Workshops and surveys highlighted three open fronts.  
First, \emph{agentic planning}—letting models reason over retrieval actions—promises better long‑horizon reasoning but raises cost and safety questions.  
Second, \emph{structured retrieval} (graphs, tables, multimodal stores) demands new embedding spaces and fusion operators.  
Third, \emph{secure and privacy‑preserving RAG} is gaining urgency, with SafeRAG prompting work on attack‑aware retrievers and watermarking of retrieved evidence.

\textbf{Outlook} Mid‑2025 results suggest that modest‑sized, security‑hardened, agent‑controlled RAG systems can rival much larger closed‑book LMs while offering provenance.  
Key challenges ahead include trusted evidence ranking at trillion‑document scale, automatic detection of retrieval‑based attacks, and seamless integration of non‑text modalities.  
Nevertheless, 2025 firmly positions RAG not merely as a booster of accuracy but as an essential framework for \emph{reliable}, \emph{updatable}, and \emph{auditable} language agents.

\section{RAG for Proprietary Data - Industry Implementation}
\label{sec:proprietary-data}

\subsection{Current Approaches}
Organizations increasingly apply RAG to \emph{private internal knowledge}, using a secure pipeline to retrieve from proprietary documents and feed them to a generative model. This often involves \emph{on-premise or VPC-hosted} vector databases, ensuring that queries and document embeddings never leave the corporate firewall
\citep{Merritt2023, IBM2023}. Enterprises store their
text in an index, and at inference time, a local embedding model transforms a user query into a dense vector to find top-$k$ relevant chunks. These chunks are appended to the user query as context for an LLM (e.g., GPT-4, or a self-hosted model). The separation of knowledge in a database reduces the risk that proprietary data leaks into the
model's parameters, but it does not fully guarantee privacy---the generator might still reveal sensitive details in the output. To mitigate these risks, some enterprise solutions incorporate \emph{access control layers} that filter out documents the user lacks
permission to see. Others experiment with \emph{secure enclaves} or homomorphic encryption to blind the retrieval operation \citep{zhao2024frag}. However, performance overhead is non-trivial. Another strategy is to \emph{fine-tune} an LLM on domain data. Although fine-tuning helps the model internalize domain nuances, it can
potentially memorize private text. RAG better suits \emph{data freshness}: the knowledge store can be updated without retraining.

\subsection{Industry Implementation Examples}
In recent years, several organizations have explored the deployment of Retrieval-Augmented Generation (RAG) systems to leverage proprietary data effectively. Notable case studies include:

\paragraph{PGA Tour's Use of RAG for Enhanced Information Retrieval}

The PGA Tour implemented a RAG system to improve information retrieval related to golf events and player statistics. By integrating their extensive proprietary data into the RAG framework, they enabled more accurate and contextually relevant responses to user queries. This approach addressed previous limitations where general AI models lacked specific domain knowledge, leading to inaccuracies in responses. The RAG system allowed the PGA Tour to provide precise information, enhancing user engagement and satisfaction \citep{wsj2024golf}.

\paragraph{Bayer's Application of RAG in Agricultural Data Management}

Bayer utilized RAG to manage and retrieve proprietary agricultural data, aiming to provide farmers with accurate and timely information. By incorporating their extensive datasets into the RAG system, Bayer enhanced the accessibility and usability of critical agricultural information. This integration facilitated better decision-making processes for farmers, leading to improved crop management and productivity \citep{wsj2024golf}.

\paragraph{Rocket Companies' Integration of RAG for Mortgage Processing}

Rocket Companies explored the use of RAG to streamline mortgage processing by integrating proprietary financial data. The RAG system enabled more efficient retrieval of relevant information, reducing processing times and improving customer experiences. By leveraging their internal data within the RAG framework, Rocket Companies enhanced the accuracy and speed of their services, demonstrating the potential of RAG in financial applications \citep{wsj2024golf}.

\paragraph{Shorenstein Properties' Implementation of RAG for Data Organization}

Shorenstein Properties adopted RAG to automate file tagging and organize proprietary data more efficiently. By integrating their internal documents into the RAG system, they improved data accessibility and management. This implementation showcased RAG's capability to handle complex data organization tasks, leading to increased operational efficiency within the company \citep{wsj2024golf}.

\paragraph{Cohere's Advancement of RAG for Source Citation}

Cohere advanced RAG technology to ensure AI systems cite their sources, facilitating human verification of the information produced. By integrating external texts such as company documents or news websites, Cohere's RAG system reduced errors known as hallucinations and provided access to current information. This development highlighted RAG's potential in enhancing the reliability and transparency of AI-generated content \citep{time2024patrick}.

\paragraph{NVIDIA's Deployment of RAG in Enterprise AI Solutions}

NVIDIA incorporated RAG into their enterprise AI solutions to connect large language models with specific information, such as proprietary customer data and authoritative research. This integration enabled more accurate and relevant responses to user queries, enhancing productivity and protecting against AI hallucinations. NVIDIA's deployment demonstrated RAG's applicability in various industries, including customer service and data management \citep{nvidia2023rag}.

\paragraph{IBM's Utilization of RAG for Domain-Specific AI Applications}

IBM employed RAG to equip models with specific information, such as proprietary customer data and authoritative research documents. This approach allowed their AI systems to incorporate up-to-date information into generated responses, improving accuracy and relevance in domain-specific applications. IBM's utilization of RAG showcased its effectiveness in enhancing AI capabilities across different sectors \citep{ibm2023rag}.

These case studies illustrate the diverse applications of RAG in leveraging proprietary data across various industries. By integrating internal datasets into RAG systems, organizations have enhanced information retrieval, improved decision-making processes, and increased operational efficiency. However, challenges such as data privacy, intellectual property concerns, and computational overhead remain areas for further research and development.

\subsection{Emergent Trends and Patterns in Industry Implementation}

\paragraph{Accuracy vs. Latency Trade-offs:} Models optimized for retrieval accuracy (e.g., FiD, WebGPT) typically have higher latency due to multiple document processing. Industry applications prioritize real-time performance (e.g., NVIDIA RAG aims for sub-second latency).

\paragraph{Scaling Beyond Wikipedia:} Early RAG models focused on Wikipedia-scale retrieval, while industry applications integrate proprietary databases and real-time web data. WebGPT and IBM Watsonx RAG retrieve from dynamic knowledge bases, offering more up-to-date responses.

\paragraph{Reducing Hallucination:} There is a shift toward factual consistency. WebGPT and IBM Watsonx RAG enforce citation mechanisms to enhance factual accuracy.

\paragraph{Model Size vs. Retrieval Efficiency:} Smaller models augmented with retrieval (e.g., Atlas, NVIDIA RAG) can outperform much larger models without retrieval (e.g., GPT-3 175B), demonstrating that knowledge retrieval reduces the need for extreme model scaling.

\paragraph{Enterprise Adaptation:} Academic models optimize for benchmark performance, whereas enterprise RAG systems prioritize real-world reliability, integration with databases, and data privacy compliance.

\section{Evaluation of RAG Systems}

The evaluation of RAG systems is multifaceted, as performance depends not only on the generative model but also on the quality of the retrieval pipeline. A robust evaluation framework must assess retrieval accuracy, answer quality, factuality, latency, and scalability. This section provides a structured overview of RAG evaluation criteria, benchmarks, and the impact of architectural design choices. Summary of this section is in Table Table~\ref{tab:rag_evaluation}

\subsection{Evaluation Dimensions and Metrics}

RAG performance is commonly assessed along the following dimensions:

\paragraph{Retrieval Accuracy.} Key metrics include:
\begin{itemize}
    \item \textbf{Recall@$k$} – Measures the proportion of queries where a relevant document appears among the top-$k$ retrieved results.
    \item \textbf{Mean Reciprocal Rank (MRR)} – Captures the average inverse rank of the first relevant document, rewarding early placement.
    \item \textbf{Mean Average Precision (MAP)} – Evaluates the quality of ranked retrieval across relevant items.
\end{itemize}

\paragraph{Generation Quality.} Evaluated using:
\begin{itemize}
    \item \textbf{Exact Match (EM)} and \textbf{F1 Score} – Common in QA tasks to measure overlap with reference answers.
    \item \textbf{BLEU}, \textbf{ROUGE} – N-gram-based measures used in summarization and long-form generation.
    \item \textbf{Faithfulness / Hallucination Rate} – Human or automated evaluations of factual consistency with retrieved sources.
\end{itemize}

\paragraph{Efficiency and Latency.} These include:
\begin{itemize}
    \item \textbf{Retrieval time}, \textbf{generation latency}, and \textbf{end-to-end response time}.
    \item \textbf{Memory and compute requirements} – Especially important for deploying RAG systems at scale.
\end{itemize}

\paragraph{Scalability.} As corpus size grows, the system’s ability to maintain retrieval quality and generation fidelity is tested. Evaluation considers:
\begin{itemize}
    \item \textbf{Index size vs. retrieval accuracy}.
    \item \textbf{Adaptability to new or evolving data without retraining}.
\end{itemize}

\subsection{Benchmarks and Datasets}

Several benchmarks are widely used to evaluate RAG systems:

\begin{itemize}
    \item \textbf{Natural Questions (NQ)}, \textbf{TriviaQA}, and \textbf{WebQuestions} for open-domain QA \citep{Lewis2020}.
    \item \textbf{FEVER} and \textbf{AveriTeC} for fact-checking, emphasizing verifiability.
    \item \textbf{KILT} benchmark suite integrates QA, dialog, slot filling, and entity linking over Wikipedia.
    \item \textbf{BEIR} evaluates retrieval across 31 zero-shot tasks in domains like biomedicine and finance.
    \item \textbf{MTRAG} targets multi-turn conversations requiring sequential retrieval and reasoning.
    \item \textbf{TREC RAG Track (2024–)} defines unified evaluation of retrieval, generation, and support quality over MS MARCO with metrics like nugget recall and citation coverage \citep{Pradeep2024}.
\end{itemize}

\subsection{Retrieval-Augmented Generation Assessment System}
RAGAS (Retrieval-Augmented Generation Assessment System) is an evaluation framework specifically designed for assessing and improving the factuality and grounding of RAG systems. Unlike conventional metrics that measure superficial linguistic overlap, RAGAS emphasizes the alignment between generated content and retrieved documents, providing explicit signals regarding factual correctness and attribution quality. By systematically measuring how well the generated outputs are supported by the retrieved evidence, RAGAS helps identify and penalize hallucinations—instances where the model generates plausible but unsupported statements. Consequently, employing RAGAS during model training or iterative fine-tuning guides RAG systems toward producing outputs firmly grounded in verifiable sources, substantially improving factual accuracy and reducing the incidence of hallucinated information.

\end{multicols}
\begin{table}[ht]
\centering
\caption{Evaluation Dimensions, Metrics, Benchmarks, and Tools for RAG Systems}
\label{tab:rag_evaluation}
\begin{tabular}{@{}lll@{}}
\toprule
\textbf{Category} & \textbf{Metric/Tool} & \textbf{Description} \\
\midrule
\multirow{3}{*}{Retrieval Accuracy} 
  & Recall@$k$ & Proportion of queries with a relevant doc in top-$k$ results \\
  & MRR & Average inverse rank of the first relevant document \\
  & MAP & Mean precision across all relevant retrieved documents \\
\midrule
\multirow{3}{*}{Generation Quality} 
  & Exact Match (EM), F1 & Measures overlap with ground-truth answers \\
  & BLEU, ROUGE & N-gram based overlap metrics for generation \\
  & Faithfulness Rate & Consistency with retrieved context (manual/automated) \\
\midrule
\multirow{2}{*}{Efficiency} 
  & Latency (retrieval/generation) & Time required for each processing step \\
  & Memory/Compute Usage & Resource demands for model operation \\
\midrule
\multirow{2}{*}{Scalability} 
  & Index Size vs.\ Accuracy & Impact of increasing corpus size on performance \\
  & Adaptability & Ability to incorporate new data without retraining \\
\midrule
\multirow{6}{*}{Benchmarks} 
  & Natural Questions, TriviaQA & Open-domain QA evaluation \\
  & FEVER, AveriTeC & Fact-checking datasets \\
  & KILT & Multi-task benchmark over Wikipedia \\
  & BEIR & 31-task zero-shot retrieval benchmark \\
  & MTRAG & Multi-turn reasoning benchmark \\
  & TREC RAG Track & Unified eval.\ of retrieval, generation, citation \\
\midrule
Tooling & RAGAS & Evaluates factual consistency and grounding \\
\bottomrule
\end{tabular}
\end{table}
\begin{multicols}{2}

\subsection{Impact of Architecture on Evaluation}

Empirical studies demonstrate that architectural components in RAG—chunking, embedding, and re-ranking—directly affect performance across tasks and benchmarks. 

Fixed-size chunks, as used in early RAG \citep{Lewis2020}, are prone to semantic fragmentation. Semantic chunking (e.g., \textit{ChunkRAG}) improves retrieval precision and answer accuracy by aligning chunk boundaries with discourse structure \citep{Singh2025ChunkRAG}.
The choice of embedding model (e.g., DPR \citep{Karpukhin2020}, HyDE \citep{Ding2024RAGsurvey}) influences Recall@$k$ and downstream generation quality. Similarly, re-ranking strategies (e.g., MonoT5) can boost EM and F1 scores by prioritizing more relevant passages.

In high-performing RAG systems, retrieval fidelity correlates strongly with factual answer quality. Thus, optimizing these architectural stages is essential for achieving competitive results on QA and generation benchmarks. 
\newpage
\section{Challenges of RAG}
This section discusses the challenges of RAG, cases of manifestation of such challenges in the selected domain of RAG application, and outlines existing solutions and the way forward.
\subsection{Technical Challenges in RAG}
\subsubsection{Retrieval Quality and Relevance}
The quality of retrieved documents significantly impacts the accuracy of RAG-generated answers. High recall and precision are critical since poor retrieval leads directly to incorrect or irrelevant answers \citep{Miao2024NephrologyRAG}. Traditional methods like BM25 are limited, often missing relevant texts or returning noisy results \citep{Chen2017WikipediaQA}. Modern neural retrievers with dense embeddings improve performance but still face issues like vocabulary mismatches, ambiguous queries, and domain-specific terminology. Specialized domain tuning, such as using legal embeddings or medical synonym expansion, can help, but maintenance of these tailored retrievers remains challenging.
Determining the optimal number of retrieved passages ($k$) is also complex. Too few passages limit evidence; too many overwhelm the model and introduce irrelevant context. Approaches like ranking retrieved passages or iterative query reformulation can improve retrieval precision, but add complexity and latency \citep{Izacard2021, Gao2024RAGSurvey}.
\subsubsection{Latency and Efficiency}
RAG inherently increases computational complexity and latency compared to standalone LLMs due to retrieval overhead, vector searches, and expanded context processing. Techniques like approximate nearest neighbor indices (e.g., FAISS, HNSW), caching, model distillation, or lightweight retrievers can reduce latency at the expense of accuracy. Integrating retrieval efficiently with large language models (LLMs) and ensuring rapid responses in real-time scenarios (e.g., customer support) remains a significant challenge \citep{Akkiraju2024FACTS}. Interestingly, using retrieval can allow smaller models to match the performance of larger models without retrieval (e.g., RETRO, Atlas), reducing model size requirements but shifting complexity to maintaining external knowledge bases and infrastructure.
\subsubsection{Integration with Large Language Models}
Integrating retrieved evidence effectively with LLMs is subtle. Models may ignore retrieved evidence, especially when internal model knowledge conflicts with external retrieved information, leading to a "tug-of-war" effect \citep{Jin2024KnowledgeConflicts}. Multiple retrieved documents might create confusion or confirmation bias if they contradict each other. Limited input lengths in transformer-based LLMs exacerbate these integration challenges by forcing truncation or summarization, potentially omitting essential context.
Fine-tuning models specifically for retrieval-augmented tasks often yields better integration than simple zero-shot prompting but introduces complexity, especially when using non-differentiable or API-based models that do not support custom training.
\subsection{System-Level Challenges}
\subsubsection{Scalability and Infrastructure}
Deploying RAG at scale requires substantial engineering to maintain large knowledge corpora and efficient retrieval indices. Systems must handle millions or billions of documents, demanding significant computational resources, efficient indexing, distributed computing infrastructure, and cost management strategies \citep{Gao2024RAGSurvey}. Efficient indexing methods, caching, and multi-tier retrieval approaches (such as cascaded retrieval) become essential at scale, especially in large deployments like web search engines.
\subsubsection{Freshness and Knowledge Updates}
One motivation for RAG is providing current information. However, continuously updating external knowledge bases and retrieval indices is challenging. Domains requiring real-time updates (e.g., finance, healthcare) demand sophisticated data pipelines for incremental updates, possibly frequent re-encoding of documents, and synchronization of retrieval indices. Delays in updates or inconsistencies between the LLM's internal knowledge and newly retrieved data can produce outdated or contradictory answers \citep{Miao2024NephrologyRAG}.
\subsubsection{Hallucination and Reliability}
While RAG reduces LLM hallucinations, it does not eliminate them completely. Models may fabricate or misattribute information if retrieval provides incomplete or partially contradictory context. Legal domain studies found that RAG significantly reduces hallucinations, but still generates errors at concerning rates \citep{Magesh2025LegalHallucinations}. Hallucinations also occur in citation generation, with models occasionally inventing nonexistent references. Strategies such as verifying outputs against retrieved sources or calibrating model confidence are needed, but no approach completely prevents hallucination.
\subsubsection{Complex Pipeline and Maintenance}
RAG systems comprise multiple components—retrievers, rerankers, indexes, and LLMs—resulting in increased complexity and potential points of failure. Maintenance includes synchronizing knowledge updates, managing access controls, orchestrating prompts, and handling multi-turn dialogues. Robust evaluation methods must assess end-to-end performance, retrieval quality, and faithfulness of model outputs to the evidence \citep{Akkiraju2024FACTS}.
\subsection{Ethical and Societal Concerns}
\subsubsection{Bias and Fairness}
RAG inherits biases from both underlying language models and retrieved external data. Biases may manifest through selective use or amplification of biased retrieved evidence, especially in historically biased domains like legal or medical information \citep{BaezaYates2018Bias,Yang2025RAGHealthcare}. Ensuring fairness involves curating inclusive datasets, using diverse retrieval results, and possibly prompting LLMs explicitly toward balanced responses.
\subsubsection{Trustworthiness and Misinformation}
The quality and reliability of retrieved sources directly impact trustworthiness. If misinformation is retrieved, the model can inadvertently disseminate false information. Even credible sources can become outdated or contextually misapplied, leading users to overly trust synthesized AI responses. Transparency in citing sources, maintaining high-quality data, and incorporating verification methods are crucial safeguards.
\subsubsection{Privacy and Security}
RAG systems handling sensitive data, especially in enterprise or healthcare contexts, raise serious privacy concerns. Risks include accidental exposure of confidential information and vulnerabilities to prompt injection attacks. Implementing rigorous access control, complying with regulations (e.g., GDPR, HIPAA), transparent data usage policies, and security testing are essential to protect privacy and prevent breaches.
\subsubsection{Accountability and Transparency}
RAG's use of sourced retrieval provides an advantage for accountability, allowing users to trace AI-generated responses back to evidence. However, inaccurate citations or improper synthesis can mislead users. Ethical deployment involves clearly attributing evidence, providing explanations on request, managing user expectations, and clearly delineating accountability—especially when RAG informs critical decisions. Transparency and accountability require ongoing evaluation, oversight, and mechanisms for user feedback and correction \citep{Magesh2025LegalHallucinations}.

\subsection{Application Domains and Case Studies}

To concretely illustrate the discussed challenges, the following section examines Retrieval-Augmented Generation (RAG) applications across three distinct domains: legal, medical, and customer support. Each domain has unique requirements influencing the design and deployment of RAG systems.

\subsubsection{Legal Domain}

Legal practice inherently involves extensive information retrieval from statutes, regulations, and case precedents, making it particularly suited for RAG applications. Commercial legal AI systems such as those provided by Westlaw and LexisNexis leverage RAG for tasks including legal research and case analysis. Accuracy and reliability are paramount; errors such as citing nonexistent cases can lead to severe professional consequences, underscored by incidents involving fabricated case citations generated by general-purpose LLMs such as ChatGPT. RAG attempts to mitigate this by ensuring assertions trace directly to verified sources.

Key challenges include:
\begin{itemize}
\item \textbf{Complex Retrieval Requirements}: Legal queries often require precise, multifaceted retrieval considering jurisdiction, topic, and specific legal doctrines. Traditional boolean-based retrieval combined with curated metadata typically addresses these needs, while neural retrieval methods struggle with contextual relevance and precision, occasionally retrieving irrelevant yet semantically similar cases \citep{CBR_RAG_Legal}.
\item \textbf{Document Length and Context Limitations}: Legal texts such as court opinions or statutes can be extensive, challenging LLM context windows. Summarizing these documents risks omitting crucial details, thereby potentially misleading practitioners. Techniques to mitigate these risks include retrieving case holdings or headnotes, though these condensed forms may omit critical nuances.

\item \textbf{Citation Standards}: Lawyers expect citations in standardized formats (e.g., Bluebook in the U.S.). AI-generated citations frequently require extensive formatting refinement and precise pinpoint references to be practically useful.

\item \textbf{Legal Reasoning and Application}: Legal analysis often involves applying retrieved legal precedents to novel scenarios. Current RAG systems primarily provide raw information, whereas nuanced legal reasoning and judgment remain beyond their current capabilities. RAG tools thus typically function as assistants rather than autonomous legal advisors.

\item \textbf{Information Freshness and Legal Updates}: Rapidly changing legal frameworks necessitate continuous updating of the retrieval corpus to maintain accuracy. Outdated citations or ignorance of recent rulings can critically undermine legal arguments, underscoring the need for dynamic, regularly updated knowledge sources.
\end{itemize}
Despite these challenges, RAG promises significant benefits by reducing research time and democratizing access to legal information. Ethically, clear boundaries must exist between supporting lawyers and unauthorized legal practice, necessitating explicit disclosures of limitations and human oversight \citep{Magesh2025LegalHallucinations}.

\subsubsection{Medical Domain}

Medical and healthcare applications of RAG involve clinical decision support, patient information summarization, and consumer-facing health advice systems. These applications commonly integrate authoritative medical databases such as PubMed and UpToDate to provide reliable, evidence-based responses.

Domain-specific challenges include:
\begin{itemize}
\item \textbf{Authoritative Source Management}: Medical RAG systems must exclusively retrieve from rigorously vetted medical literature, clinical guidelines, and patient databases, necessitating careful curation and precise summarization to avoid misinterpretation or oversimplification.
\item \textbf{Rapid Evolution of Medical Knowledge}: Rapid developments, highlighted during events such as the COVID-19 pandemic, require systems to continuously integrate evolving medical evidence and guidelines. Misrepresentations due to outdated or disproven studies pose significant risks, emphasizing the importance of dynamically updated retrieval sources and structured medical knowledge bases \citep{Gilbert2024MedCurator}.

\item \textbf{Diagnostic Reasoning Limitations}: Medical diagnostics often require complex reasoning rather than straightforward factual retrieval. Current RAG implementations perform best on factual inquiries (e.g., drug dosages, treatment guidelines), while open-ended diagnostic queries involving symptom analysis remain challenging and risk inappropriate advice.

\item \textbf{Privacy and Patient Data Security}: Ensuring patient privacy when integrating patient-specific records into RAG systems demands robust data handling protocols to avoid breaches or inappropriate data mixing.

\item \textbf{Bias in Medical Data}: Historical underrepresentation in clinical trials introduces biases that RAG systems may perpetuate if not addressed proactively through diverse and balanced source curation \citep{Yang2025RAGHealthcare}.

\item \textbf{Regulatory Compliance and Liability}: RAG systems providing medical advice may fall under regulatory oversight such as FDA guidelines, requiring explicit disclaimers and positioning as decision-support tools rather than autonomous advisors.
\end{itemize}

The medical domain emphasizes the necessity for high factual accuracy, ethical transparency, and rigorous validation processes comparable to clinical trials. Successful deployment hinges on clear delineation between supportive advisory roles and direct patient intervention \citep{Nori2023MedicalGPT}.

\subsubsection{Customer Support and Knowledge Bases}

RAG systems increasingly automate customer support tasks by retrieving information from internal knowledge bases, FAQs, and troubleshooting guides. While typically lower stakes compared to legal or medical domains, customer support RAG applications encounter distinct practical considerations:

\begin{itemize}
\item \textbf{Flexible Query Interpretation}: Support queries are often informal or ambiguous. Semantic retrieval effectively handles varied user inputs but requires precise handling to avoid irrelevant answers.
\item \textbf{Multi-turn Interaction Management}: Effective customer support involves maintaining conversation context and dynamically retrieving information based on prior user interactions. This requires complex dialogue management integrated with retrieval processes.

\item \textbf{Personalization and Privacy Considerations}: Customer-specific information (e.g., orders, account details) necessitates secure, privacy-aware integration within retrieval processes to avoid data misuse.

\item \textbf{Knowledge Base Quality and Maintenance}: The utility of RAG systems is directly tied to the quality and currency of support documentation, highlighting the critical need for continual updating and management of the knowledge corpus.

\item \textbf{User Experience and Tone}: Customer support demands empathetic, conversational responses that match user expectations. Ethical transparency in disclosing the AI nature of interactions is essential.

\item \textbf{Escalation and Handling Failures}: Systems must effectively identify limits of their capabilities, promptly escalating unresolved issues to human support to maintain customer satisfaction.

\item \textbf{Metrics for Evaluation}: Success in customer support is evaluated through resolution rates and customer satisfaction metrics, requiring responses that not only provide factual accuracy but practical utility.
\end{itemize}

Ethical considerations include transparency, data security, job impact, and fairness. RAG implementations also provide valuable feedback for continual improvement of internal support documentation based on real-time user interactions, thereby enhancing both system performance and documentation quality.

\subsection{Existing and Potential Solutions}

\subsubsection{Retrieval Quality}
Maintaining high retrieval relevance is critical for effective RAG. Strategies to improve retrieval quality include domain-adaptive training, advanced encoders, and query reformulation methods to address vocabulary mismatches \citep{siriwardhana2023improving}. Employing reranking models further boosts relevance by re-scoring initial retrieval results with deeper contextual analysis, enhancing accuracy at the expense of additional computation \citep{An2025}. Iterative retrieval and chain-of-thought reasoning represent future directions, breaking down complex queries into simpler sub-queries, thus ensuring relevant information retrieval at each reasoning step \citep{Trivedi2023}.

\subsubsection{Latency}
RAG systems introduce latency due to retrieval processes. Solutions include using efficient nearest-neighbor search structures, such as HNSW graphs, which significantly speed up similarity searches \citep{Malkov2018}. Caching mechanisms, including multi-level and approximate embedding caches (e.g., RAGCache and Proximity cache), enable reuse of previously retrieved information, drastically reducing retrieval time \citep{Jin2024,Bergman2025}. Adaptive retrieval methods dynamically balance retrieval complexity based on query difficulty, optimizing overall throughput and reducing latency.

\subsubsection{Model Integration}
Effective integration between retrieval and generation models remains essential. Methods include joint end-to-end training of retrievers and generators, enhancing mutual compatibility and performance \citep{Lewis2020}. Architectural integration techniques, such as RETRO's cross-attention mechanism, dynamically incorporate retrieved facts during generation \citep{Borgeaud2022}. Alternatively, prompt-based integration treats LLMs as black-boxes, conditioning on retrieved documents without architectural modifications. Future hybrid approaches involving reinforcement learning and selective retrieval aim to optimize when and how external knowledge is incorporated into generation processes.

\subsubsection{Hallucination}
Reducing factual hallucinations remains a key focus. RAG inherently mitigates hallucinations by grounding outputs in retrieved evidence \citep{Shuster2021}. Training models to penalize ungrounded assertions and iterative retrieval within reasoning processes further enhance accuracy \citep{Trivedi2023}. Self-check mechanisms (Self-RAG), where models critique and revise their outputs against retrieval results, significantly reduce hallucinated content \citep{Asai2024}. External verification and fact-checking modules complement internal methods, collectively ensuring high factual reliability. For instance, RAG systems to cite sources significantly enhance their reliability by directly linking generated information to supporting evidence. This citation capability plays a crucial role in mitigating the common issue of hallucination, where generative models produce plausible yet inaccurate or fabricated information. By explicitly associating each factual statement with retrieved documents, RAG systems encourage transparency and verifiability, enabling users and downstream processes to quickly assess the accuracy and provenance of claims. Moreover, requiring the model to cite sources during generation inherently promotes grounding outputs in verified data, further reducing the risk of generating unsupported statements \citep{Shuster2021}. Thus, citation functionality not only enhances user trust but also fosters more disciplined, factually accurate generation, substantially decreasing the likelihood of hallucinated outputs.

\subsubsection{Scalability}
Scalability challenges arise as knowledge corpora expand. Advanced indexing, distributed retrieval, and approximate nearest neighbor techniques facilitate efficient handling of large-scale knowledge bases \citep{Malkov2018}. Selective indexing and corpus curation, combined with infrastructure improvements like caching and parallel retrieval, allow RAG systems to scale to massive knowledge repositories. Research indicates that moderate-sized models augmented with large external corpora can outperform significantly larger standalone models, suggesting parameter efficiency advantages \citep{Borgeaud2022}.

\subsubsection{Knowledge Freshness}
Rapidly evolving information necessitates regularly updated knowledge bases. RAG systems can efficiently maintain knowledge freshness through incremental updates and selective retrieval methods without requiring frequent retraining \citep{He2025}. Integrating live search APIs and hybrid retrieval methods ensure real-time information retrieval, addressing dynamic knowledge demands \citep{Gao2024RAGSurvey}. Continuous updates and user-feedback integration support lifelong learning and timely information access.

\subsubsection{Bias}
Addressing bias in RAG involves curating balanced knowledge sources, employing diversification techniques in retrieval, and adjusting retriever embeddings to counteract inherent biases \citep{Kim2025}. Prompts and model training that encourage balanced representation, along with transparency in source attribution, further mitigate bias propagation. This multi-faceted approach helps minimize biases in RAG outputs.

\subsubsection{Misinformation}
Combating misinformation involves preventive measures like curating trustworthy knowledge sources and reactive verification through stance classifiers and credibility assessments \citep{Pan2023}. Models employing vigilant prompting, cross-verification with multiple retrieved documents, and external fact-checking modules enhance reliability and truthfulness. Robustness against adversarial misinformation insertion through continuous monitoring and data validation further strengthens RAG systems, ensuring accurate information dissemination.

\section{Discussion and Future Direction}
\label{sec:discussion}

\subsection{Synthesis of Findings}
\label{subsec:synthesis}

Retrieval-Augmented Generation (RAG) has emerged as a paradigm shift in AI, enabling language models to dynamically retrieve and incorporate external knowledge. Studies confirm that RAG-based models significantly outperform purely parametric generative models, achieving state-of-the-art results on knowledge-intensive NLP tasks \citep{Lewis2020, Karpukhin2020}. Lewis et al. \citep{Lewis2020} demonstrated that RAG surpasses fine-tuned BART on open-domain question answering (QA), generating more specific and factual responses.

The historical development of RAG is rooted in the evolution of open-domain QA architectures. Early approaches like DrQA \citep{Chen2017WikipediaQA} relied on sparse lexical retrieval, whereas modern RAG implementations integrate differentiable dense retrievers such as Dense Passage Retrieval (DPR) \citep{Karpukhin2020}. By conditioning generation on retrieved evidence, RAG mitigates issues like hallucinations and stale knowledge \citep{Izacard2021}. The comparative analysis of different RAG architectures reveals trade-offs in retrieval effectiveness, generative fluency, and computational cost.

Despite continuous improvements, challenges remain. RAG's dependency on retrieval quality makes it vulnerable to retrieval failures. Errors in retrieving relevant documents lead to incorrect outputs. Additionally, integrating multiple retrieved passages introduces challenges in fusion mechanisms, sometimes resulting in contradictory evidence or increased latency. The need for robust retrieval strategies and enhanced fusion methods remains a critical research direction.

\subsection{Implications for Proprietary Data}
\label{subsec:proprietary_data}

A key development in RAG is its application to enterprise AI, enabling access to proprietary data without embedding sensitive information into model parameters \citep{Izacard2022}. This architecture supports data privacy and ensures outputs remain grounded in up-to-date knowledge \citep{aws2023rag}. Unlike traditional LLMs, which require retraining to update internal knowledge, RAG allows retrieval components to be independently refreshed \citep{aws2023rag}.

However, enterprise RAG deployments introduce new security concerns. The retriever could inadvertently expose confidential information if strict access controls are not enforced. Organizations must implement robust authentication mechanisms, ensuring that retrieved documents align with user permissions. Privacy-preserving techniques such as encrypted search indices and federated retrieval \citep{federatedlearning2021} are promising solutions to mitigate risks.

The ability of RAG to function over proprietary knowledge bases has made it a preferred choice for industries handling sensitive information, including finance, healthcare, and legal sectors. As enterprises scale RAG systems, optimizing retrieval latency and ensuring regulatory compliance will be paramount.

\subsection{Future Research Directions}
Key avenues for advancing RAG include:
\begin{itemize}
\item \textbf{Multi-hop retrieval}: RAG pipelines typically retrieve single passages, but many queries require reasoning over multiple documents. Standard RAG models struggle with multi-hop questions \citep{Tang2024}, motivating research on multi-step retrieval and reasoning (e.g., chained or iterative RAG methods).
\item \textbf{Secure and privacy-aware retrieval}: As RAG is applied to sensitive data, privacy becomes critical. Recent work proposes encrypting knowledge bases and embeddings to prevent unauthorized access while preserving performance \citep{Zhou2025}, and integrating differential privacy and secure computation to guard against information leakage \citep{Kandula2025}.
\item \textbf{Multimodal and agentic RAG}: Extending RAG beyond text can leverage rich context. Multimodal RAG (MRAG) systems incorporate images, video, or other modalities into retrieval and generation, reducing hallucinations and improving performance on visual queries \citep{Mei2025}.  Separately, \emph{Agentic} RAG embeds autonomous agents into the retrieval pipeline, enabling dynamic planning, tool use, and multi-step reasoning in the loop \citep{singh2025agentic}.
\item \textbf{Structured knowledge integration}: Traditional RAG relies on vector search over unstructured text, which can miss complex relationships. Incorporating structured knowledge (e.g., knowledge graphs) can improve semantic understanding. Surveys and systems demonstrate that linking RAG to knowledge graphs yields more accurate and coherent answers \citep{Chen2025,Guo2024}.
\item \textbf{Real-time or streaming retrieval}: Many applications require RAG to operate on live data streams. For example, systems like \emph{StreamingRAG} build evolving knowledge graphs from streaming inputs to provide timely context updates, achieving much faster throughput and higher temporal relevance than static RAG \citep{Sankaradas2025}.
\end{itemize}

\subsection{Emerging Practical Applications}
RAG is finding use in a range of new applications:
\begin{itemize}
\item \textbf{Enterprise search}: Companies employ RAG to query internal document repositories. By building RAG pipelines around enterprise vector databases, organizations can retrieve and synthesize answers from private knowledge with strong performance, scalability, and security guarantees \citep{Harvey2025}.
\item \textbf{Real-time assistants}: Personal or domain-specific AI assistants can use RAG to retrieve current information (e.g., news, stock prices, weather) on the fly. The RAG framework naturally supports grounding LLM responses in real-time data, improving accuracy and user trust \citep{IBM2023}.
\item \textbf{Fact-checking}: Integrating RAG into fact-checking pipelines can automate evidence retrieval. Recent systems use RAG-based reasoning (e.g., chain-of-RAG) to fetch and synthesize evidence for claims, improving veracity predictions in political and news domains \citep{AbdulKhaliq2024}.
\item \textbf{Conversational agents}: Chatbots and customer-service agents increasingly leverage RAG to provide informed responses. For instance, a RAG-based response system for contact centers has shown higher accuracy and relevance than traditional models \citep{Veturi2024}, helping agents answer queries with up-to-date information.
\item \textbf{Low-resource QA}: In domains with scarce curated data, RAG can bootstrap question answering by harnessing alternative corpora. A two-layer RAG system using social-media content answered medical questions accurately on modest hardware, demonstrating viability in low-resource settings \citep{Das2025}.
\end{itemize}

\section{Conclusion}
\label{sec:conclusion}

Retrieval-Augmented Generation (RAG) represents a fundamental advancement in AI, bridging retrieval and generation for improved factuality and adaptability. This review highlights the evolution of RAG from early retrieve-and-read systems to sophisticated architectures integrating neural retrievers and sequence-to-sequence generators. RAG has demonstrated significant benefits across open-domain QA, enterprise applications, and AI-powered search.

While RAG enhances generative AI with dynamic knowledge retrieval, several challenges remain. Ensuring high-quality retrieval, handling conflicting retrieved evidence, and scaling retrieval mechanisms for large knowledge bases require further research. Additionally, security considerations in proprietary deployments necessitate privacy-preserving retrieval strategies.

Future research should focus on optimizing retrieval efficiency, refining document fusion strategies, and developing robust evaluation metrics for retrieval-augmented generation. The continued convergence of information retrieval and AI-powered text generation will define the next generation of intelligent assistants, transforming how users interact with digital knowledge.

\section*{Acknowledgments}
This work (J.A and A.B) was supported by the University of Tennessee startup funding. The authors acknowledge the use of facilities and instrumentation at the UT Knoxville Institute for Advanced Materials and Manufacturing (IAMM) supported in part by the National Science Foundation Materials Research Science and Engineering Center program through the UT Knoxville Center for Advanced Materials and Manufacturing (DMR-2309083). We extend our gratitude to the faculty and staff of UT-ORII for their invaluable support.

\section*{Conflict of Interest}
The authors confirm there is no conflict of interest.

\newpage
\bibliographystyle{plain}
\bibliography{references}

\begin{thebibliography}{100}

\bibitem{AbdulKhaliq2024}
Mohammed AbdulKhaliq, Paul Yu-Chun Chang, Mingyang Ma, Bernhard Pflugfelder, and Filip Miletic.
\newblock Ragar, your falsehood radar: Rag-augmented reasoning for political fact-checking using multimodal llms.
\newblock {\em arXiv preprint arXiv:2404.12065}, 2024.

\bibitem{CBR_RAG_Legal}
Mark Agatonovic, Anastasia Shimorina, et~al.
\newblock Cbr-rag: Case-based reasoning for retrieval augmented generation in llms for legal question answering.
\newblock In {\em Proceedings of the 2024 International Conference on Artificial Intelligence and Law (ICAIL)}, 2024 (to appear).

\bibitem{Akkiraju2024FACTS}
Rama Akkiraju, Anbang Xu, Deepak Bora, Tan Yu, Lu~An, Vishal Seth, Aaditya Shukla, et~al.
\newblock {FACTS} about building retrieval augmented generation-based chatbots.
\newblock arXiv preprint arXiv:2407.07858, 2024.

\bibitem{An2025}
Yuwei An, Yihua Cheng, Seo~Jin Park, and Junchen Jiang.
\newblock Hyperrag: Enhancing quality-efficiency tradeoffs in retrieval-augmented generation with reranker kv-cache reuse.
\newblock arXiv preprint arXiv:2504.02921, 2025.

\bibitem{asai2023selfrag}
Akari Asai, Zeqiu Wu, Yizhong Wang, Avirup Sil, and Hannaneh Hajishirzi.
\newblock Self-rag: Learning to retrieve, generate, and critique through self-reflection.
\newblock {\em arXiv preprint arXiv:2310.11511}, 2023.

\bibitem{Asai2024}
Akari Asai, Zeqiu Wu, Yizhong Wang, Avirup Sil, and Hannaneh Hajishirzi.
\newblock Self-rag: Learning to retrieve, generate, and critique through self-reflection.
\newblock In {\em Proceedings of the Twelfth International Conference on Learning Representations (ICLR)}, 2024.

\bibitem{BaezaYates2018Bias}
Ricardo Baeza-Yates.
\newblock Bias on the web.
\newblock {\em Communications of the ACM}, 61(6):54--61, 2018.

\bibitem{Bergman2025}
Shai Bergman, Zhang Ji, Anne-Marie Kermarrec, Diana Petrescu, Rafael Pires, Mathis Randl, and Martijn de~Vos.
\newblock Leveraging approximate caching for faster retrieval-augmented generation.
\newblock In {\em Proceedings of the 5th Workshop on Machine Learning and Systems (EuroMLSys)}, 2025.

\bibitem{borgeaud2022retro}
Sebastian Borgeaud, Arthur Mensch, Jordan Hoffmann, Trevor Cai, Eliza Rutherford, Katie Millican, George van~den Driessche, Jean Lespiau, Bogdan Damoc, Aidan Clark, et~al.
\newblock Improving language models by retrieving from trillions of tokens.
\newblock {\em arXiv preprint arXiv:2112.04426}, 2022.

\bibitem{Borgeaud2022}
Sebastian Borgeaud, Arthur Mensch, Jordan Hoffmann, Trevor Cai, Eliza Rutherford, Katie Millican, George van~den Driessche, Jean-Baptiste Lespiau, Bogdan Damoc, Aidan Clark, Diego de~Las~Casas, Aurelia Guy, Jacob Menick, Roman Ring, Tom Hennigan, Saffron Huang, Loren Maggiore, Chris Jones, Albin Cassirer, Andy Brock, Michela Paganini, Geoffrey Irving, Oriol Vinyals, Simon Osindero, Karen Simonyan, Jack~W. Rae, Erich Elsen, and Laurent Sifre.
\newblock Improving language models by retrieving from trillions of tokens.
\newblock In {\em Proceedings of the 39th International Conference on Machine Learning (ICML)}. PMLR, 2022.

\bibitem{wsj2024golf}
Isabelle Bousquette.
\newblock Ai doesn't know much about golf. or farming. or mortgages. or ...
\newblock {\em The Wall Street Journal}, 2024.

\bibitem{Chen2017WikipediaQA}
Danqi Chen, Adam Fisch, Jason Weston, and Antoine Bordes.
\newblock Reading wikipedia to answer open-domain questions.
\newblock In {\em Proceedings of the 55th Annual Meeting of the Association for Computational Linguistics (ACL)}, pages 1870--1879, 2017.

\bibitem{chen2024benchmarking}
Jiawei Chen, Hongyu Lin, Xianpei Han, and Le~Sun.
\newblock Benchmarking large language models in retrieval-augmented generation.
\newblock In {\em Proceedings of the AAAI Conference on Artificial Intelligence}, volume~38, pages 17754--17762, 2024.

\bibitem{Chen2025}
Ruixi Chen.
\newblock Retrieval-augmented generation with knowledge graphs: A survey.
\newblock {\em arXiv preprint arXiv:2511.00000}, 2025.
\newblock CSUC 2025 submission.

\bibitem{chen2020dialogue}
Xiuyi Chen, Fandong Meng, Peng Li, Feilong Chen, Shuang Xu, Bo~Xu, and Jie Zhou.
\newblock Bridging the gap between prior and posterior knowledge selection for knowledge-grounded dialogue generation.
\newblock In {\em EMNLP}, 2020.

\bibitem{cheng2023lift}
Xin Cheng, Di~Luo, Xiuying Chen, Lemao Liu, Dongyan Zhao, and Rui Yan.
\newblock Lift yourself up: Retrieval-augmented text generation with self-memory.
\newblock {\em Advances in Neural Information Processing Systems}, 36:43780--43799, 2023.

\bibitem{Das2025}
Sudeshna Das, Yao Ge, Yuting Guo, Swati Rajwal, JaMor Hairston, Jeanne Powell, Drew Walker, Snigdha Peddireddy, Sahithi Lakamana, Selen Bozkurt, Matthew Reyna, Reza Sameni, Yunyu Xiao, Sangmi Kim, Rasheeta Chandler, Natalie Hernandez, Danielle Mowery, Rachel Wightman, Jennifer Love, Anthony Spadaro, Jeanmarie Perrone, and Abeed Sarker.
\newblock Two-layer retrieval-augmented generation framework for low-resource medical question answering using reddit data: Proof-of-concept study.
\newblock {\em J. Med. Internet Res.}, 27:e66220, 2025.

\bibitem{Dinan2019}
Emily Dinan, Stephen Roller, Kurt Shuster, Angela Fan, Michael Auli, and Jason Weston.
\newblock Wizard of wikipedia: Knowledge-powered conversational agents.
\newblock In {\em Proceedings of the International Conference on Learning Representations (ICLR)}, 2019.

\bibitem{Ding2024RAGsurvey}
Yujuan Ding, Wenqi Fan, Liangbo Ning, Shijie Wang, Hengyun Li, Dawei Yin, Tat-Seng Chua, and Qing Li.
\newblock A survey on rag meets llms: Towards retrieval-augmented large language models.
\newblock {\em arXiv preprint arXiv:2405.06211}, 2024.

\bibitem{eibich2024arag}
Matou{\v s} Eibich, Shivay Nagpal, and Alexander Fred-Ojala.
\newblock {ARAGOG}: Advanced {RAG} output grading.
\newblock {\em arXiv preprint arXiv:2404.01037}, 2024.

\bibitem{Gao2024RAGSurvey}
Yunfan Gao, Yun Xiong, Xinyu Gao, Kangxiang Jia, Jinliu Pan, Yuxi Bi, Yi~Dai, Jiawei Sun, Meng Wang, and Haofen Wang.
\newblock Retrieval-augmented generation for large language models: A survey.
\newblock arXiv preprint arXiv:2312.10997, 2024.

\bibitem{Gilbert2024MedCurator}
Stephen Gilbert, Jakob~N. Kather, and Aidan Hogan.
\newblock Augmented non-hallucinating large language models as medical information curators.
\newblock {\em NPJ Digital Medicine}, 7(1):100030, 2024.

\bibitem{golatkar2024cpr}
Aditya Golatkar, Alessandro Achille, Luca Zancato, Yu-Xiang Wang, Ashwin Swaminathan, and Stefano Soatto.
\newblock Cpr: Retrieval augmented generation for copyright protection.
\newblock In {\em Proceedings of the IEEE/CVF Conference on Computer Vision and Pattern Recognition}, pages 12374--12384, 2024.

\bibitem{Guo2024}
Xiaoming Guo, Shengting Cao, Shenglin Li, Qi~Yin, and Cien Li.
\newblock Structugraphrag: Structured document-informed knowledge graphs for retrieval-augmented generation.
\newblock In {\em AAAI Spring Symposium on Generative AI for Social Science Research}, 2024.

\bibitem{guu2020realm}
Kelvin Guu, Kenton Lee, Zora Tung, Panupong Pasupat, and Ming-Wei Chang.
\newblock {REALM}: Retrieval-augmented language model pre-training.
\newblock In {\em Proceedings of the 37th International Conference on Machine Learning (ICML)}, pages 3929--3938, 2020.

\bibitem{Guu2020}
Kelvin Guu, Kenton Lee, Zora Tung, Panupong Pasupat, and Ming-Wei Chang.
\newblock Retrieval-augmented language model pre-training.
\newblock In {\em Proceedings of the 37th International Conference on Machine Learning (ICML)}, volume 119, pages 3929--3938, Online, 2020. PMLR.

\bibitem{hadi2023survey}
Muhammad~Usman Hadi, Rizwan Qureshi, Abbas Shah, Muhammad Irfan, Anas Zafar, Muhammad~Bilal Shaikh, Naveed Akhtar, Jia Wu, Seyedali Mirjalili, et~al.
\newblock A survey on large language models: Applications, challenges, limitations, and practical usage.
\newblock {\em Authorea Preprints}, 3, 2023.

\bibitem{han2025ragvgraph}
Haoyu Han, Harry Shomer, Yu~Wang, Yongjia Lei, Kai Guo, Zhigang Hua, Bo~Long, Hui Liu, and Jiliang Tang.
\newblock Rag vs.\ graphrag: A systematic evaluation and key insights.
\newblock {\em arXiv preprint arXiv:2502.11371}, 2025.

\bibitem{han2025graphrag}
Haoyu Han, Yu~Wang, Harry Shomer, Kai Guo, Jiayuan Ding, Yongjia Lei, Mahantesh Halappanavar, Ryan~A. Rossi, Subhabrata Mukherjee, Xianfeng Tang, Bo~Long, Tong Zhao, Neil Shah, Yinglong Xia, and Jiliang Tang.
\newblock Retrieval-augmented generation with graphs (graphrag).
\newblock {\em arXiv preprint arXiv:2501.00309}, 2025.

\bibitem{He2025}
Guoxiu He, Xin Song, and Aixin Sun.
\newblock Knowledge updating? no more model editing! just selective contextual reasoning.
\newblock {\em Journal of the ACM}, 2025.
\newblock to appear.

\bibitem{he2024g}
Xiaoxin He, Yijun Tian, Yifei Sun, Nitesh Chawla, Thomas Laurent, Yann LeCun, Xavier Bresson, and Bryan Hooi.
\newblock G-retriever: Retrieval-augmented generation for textual graph understanding and question answering.
\newblock {\em Advances in Neural Information Processing Systems}, 37:132876--132907, 2024.

\bibitem{ibm2023rag}
IBM.
\newblock What is rag (retrieval augmented generation)?
\newblock {\em IBM}, 2023.

\bibitem{IBM2023}
{IBM Research}.
\newblock What is retrieval-augmented generation?
\newblock IBM Research Blog, 22 August 2023, 2023.
\newblock \url{https://research.ibm.com/blog/retrieval-augmented-generation-RAG}.

\bibitem{Izacard2021}
Gautier Izacard and Edouard Grave.
\newblock Leveraging passage retrieval with generative models for open domain question answering.
\newblock In {\em Proceedings of the 16th Conference of the European Chapter of the Association for Computational Linguistics (EACL)}, pages 874--880, 2021.

\bibitem{Izacard2022}
Gautier Izacard, Patrick Lewis, Maria Lomeli, Lucas Hosseini, Fabio Petroni, Timo Schick, Jane Dwivedi-Yu, Armand Joulin, Sebastian Riedel, and Edouard Grave.
\newblock Atlas: Few-shot learning with retrieval augmented language models.
\newblock {\em arXiv preprint arXiv:2208.03299}, 2022.

\bibitem{izacard2023atlas}
Gautier Izacard, Patrick Lewis, Maria Lomeli, Lucas Hosseini, Fabio Petroni, Timo Schick, Jane Dwivedi-Yu, Armand Joulin, Sebastian Riedel, and Edouard Grave.
\newblock {A}tlas: Few-shot learning with retrieval augmented language models.
\newblock {\em Journal of Machine Learning Research}, 24:1--43, 2023.

\bibitem{izacard2022contriever}
Gautier Izacard, Xin Wan, Christoph B{\"o}hm, Kazuma Irie, Nathanael Sch{\"a}rli, and Sebastian Riedel.
\newblock Unsupervised dense information retrieval with contrastive learning.
\newblock {\em Transactions of the Association for Computational Linguistics}, 2022.
\newblock arXiv:2201.10672.

\bibitem{jiang-etal-2023-active}
Zhengbao Jiang, Frank Xu, Luyu Gao, Zhiqing Sun, Qian Liu, Jane Dwivedi-Yu, Yiming Yang, Jamie Callan, and Graham Neubig.
\newblock Active retrieval augmented generation.
\newblock In Houda Bouamor, Juan Pino, and Kalika Bali, editors, {\em Proceedings of the 2023 Conference on Empirical Methods in Natural Language Processing}, pages 7969--7992, Singapore, December 2023. Association for Computational Linguistics.

\bibitem{jin2024long}
Bowen Jin, Jinsung Yoon, Jiawei Han, and Sercan~O Arik.
\newblock Long-context llms meet rag: Overcoming challenges for long inputs in rag.
\newblock In {\em The Thirteenth International Conference on Learning Representations}, 2024.

\bibitem{Jin2024}
Chao Jin, Zili Zhang, Xuanlin Jiang, Fangyue Liu, Xin Liu, Xuanzhe Liu, and Xin Jin.
\newblock Ragcache: Efficient knowledge caching for retrieval-augmented generation.
\newblock arXiv preprint arXiv:2404.12457, 2024.

\bibitem{Jin2024KnowledgeConflicts}
Zhuoran Jin, Pengfei Cao, Yubo Chen, Kang Liu, Xiaojian Jiang, Jiexin Xu, Qiuxia Li, and Jun Zhao.
\newblock Tug-of-war between knowledge: Exploring and resolving knowledge conflicts in retrieval-augmented language models.
\newblock In {\em Proceedings of the 2024 Joint Conference on Computational Language Learning and Language Resources and Evaluation (LREC-COLING)}, 2024.

\bibitem{Kandula2025}
Sheshananda~Reddy Kandula.
\newblock Securing retrieval-augmented generation: Privacy risks and mitigation strategies.
\newblock {\em SSRN Electronic Journal}, 2025.

\bibitem{kang2023surge}
Minki Kang, Jin~Myung Kwak, Jinheon Baek, and Sung~Ju Hwang.
\newblock Knowledge graph-augmented language models for knowledge-grounded dialogue generation (surge).
\newblock {\em arXiv preprint arXiv:2305.18846}, 2023.

\bibitem{karpukhin2020dpr}
Vladimir Karpukhin, Barlas Oguz, Sewon Min, Patrick Lewis, Ledell Wu, Sergey Edunov, Danqi Chen, and Wen-tau Yih.
\newblock Dense passage retrieval for open-domain question answering.
\newblock In {\em Proceedings of the 2020 Conference on Empirical Methods in Natural Language Processing (EMNLP)}, pages 6769--6781, 2020.

\bibitem{Karpukhin2020}
Vladimir Karpukhin, Barlas Oguz, Sewon Min, Patrick S.~H. Lewis, Ledell Wu, Sergey Edunov, Danqi Chen, and Wen-tau Yih.
\newblock Dense passage retrieval for open-domain question answering.
\newblock In {\em Proceedings of the 2020 Conference on Empirical Methods in Natural Language Processing (EMNLP)}, pages 6769--6781, Online, 2020. Association for Computational Linguistics.

\bibitem{Kim2025}
Taeyoun Kim, Jacob Springer, Aditi Raghunathan, and Maarten Sap.
\newblock Mitigating bias in rag: Controlling the embedder.
\newblock arXiv preprint arXiv:2502.17390, 2025.

\bibitem{kumari2023dialog}
Lilly Kumari, Usama~Bin Shafqat, and Nikhil Sarda.
\newblock Retrieval-augmented generation for dialog modeling.
\newblock In {\em Proceedings of the 37th International Conference on Neural Information Processing Systems (NeurIPS 2023) Workshops}, 2023.

\bibitem{Lee2019}
Kenton Lee, Ming-Wei Chang, and Kristina Toutanova.
\newblock Latent retrieval for weakly supervised open domain question answering.
\newblock In {\em Proceedings of the 57th Annual Meeting of the Association for Computational Linguistics (ACL)}, pages 6086--6096, Florence, Italy, 2019. Association for Computational Linguistics.

\bibitem{lewis2020marge}
Mike Lewis, Marjan Ghazvininejad, Gargi Ghosh, Armen Aghajanyan, Sida Wang, and Luke Zettlemoyer.
\newblock Pre-training via paraphrasing.
\newblock {\em arXiv:2006.15020}, 2020.

\bibitem{time2024patrick}
Patrick Lewis.
\newblock Time100 ai 2024.
\newblock {\em Time}, 2024.

\bibitem{lewis2020retrieval}
Patrick Lewis, Ethan Perez, Aleksandra Piktus, Fabio Petroni, Vladimir Karpukhin, Naman Goyal, Heinrich K{\"u}ttler, Mike Lewis, Wen-tau Yih, Tim Rockt{\"a}schel, Sebastian Riedel, et~al.
\newblock Retrieval-augmented generation for knowledge-intensive nlp tasks.
\newblock In {\em Advances in Neural Information Processing Systems 33 (NeurIPS)}, pages 9459--9474, 2020.

\bibitem{Lewis2020}
Patrick Lewis, Ethan Perez, Aleksandra Piktus, Fabio Petroni, Vladimir Karpukhin, Naman Goyal, Heinrich K{\"u}ttler, Mike Lewis, Wen-tau Yih, Tim Rockt{\"a}schel, Sebastian Riedel, and Douwe Kiela.
\newblock Retrieval-augmented generation for knowledge-intensive {NLP} tasks.
\newblock In {\em Advances in Neural Information Processing Systems (NeurIPS)}, volume~33, pages 9459--9474. Curran Associates, Inc., 2020.

\bibitem{li2024omnsearch}
Yangning Li, Yinghui Li, Xinyu Wang, Yong Jiang, Zhen Zhang, Xinran Zheng, Hui Wang, Hai-Tao Zheng, Pengjun Xie, Philip~S. Yu, and Jingren Zhou.
\newblock Benchmarking multimodal retrieval augmented generation with dynamic {VQA} dataset and self-adaptive planning agent.
\newblock {\em arXiv preprint arXiv:2411.02937}, 2024.

\bibitem{li2024rag_vs_long}
Zhen Li, Cheng Li, Ming Zhang, Qiaozhu Mei, and Michael Bendersky.
\newblock Retrieval augmented generation or long-context {LLMs}? a comprehensive study and hybrid approach.
\newblock {\em arXiv preprint arXiv:2407.16833}, 2024.

\bibitem{liang2025saferag}
Xun Liang, Simin Niu, Zhiyu Li, Sensen Zhang, Hanyu Wang, Feiyu Xiong, Jason~Zhaoxin Fan, Bo~Tang, Shichao Song, Mengwei Wang, and Jiawei Yang.
\newblock Saferag: Benchmarking security in retrieval-augmented generation of large language models.
\newblock {\em arXiv preprint arXiv:2501.18636}, 2025.

\bibitem{Magesh2025LegalHallucinations}
Varun Magesh, Faiz Surani, Matthew Dahl, Mirac Suzgun, Christopher~D. Manning, and Daniel~E. Ho.
\newblock Hallucination-free? assessing the reliability of leading ai legal research tools.
\newblock {\em Journal of Empirical Legal Studies}, 0(1):1--27, 2025.
\newblock Early View, https://doi.org/10.1111/jels.12413.

\bibitem{Malkov2018}
Yu.~A. Malkov and D.~A. Yashunin.
\newblock Efficient and robust approximate nearest neighbor search using hierarchical navigable small world graphs.
\newblock {\em IEEE Transactions on Pattern Analysis and Machine Intelligence}, 40(9):2225--2237, 2018.

\bibitem{federatedlearning2021}
Priyanka~Mary Mammen.
\newblock Federated learning: Opportunities and challenges.
\newblock {\em arXiv preprint arXiv:2101.05428}, 2021.
\newblock Accessed: 2025-03-19.

\bibitem{Mei2025}
Lang Mei, Siyu Mo, Zhihan Yang, and Chong Chen.
\newblock A survey of multimodal retrieval-augmented generation.
\newblock {\em arXiv preprint arXiv:2504.08748}, 2025.

\bibitem{Merritt2023}
Rick Merritt.
\newblock What is retrieval-augmented generation aka {RAG}?
\newblock NVIDIA Blog, 15 November 2023, 2023.
\newblock \url{https://blogs.nvidia.com/blog/what-is-retrieval-augmented-generation/}.

\bibitem{Miao2024NephrologyRAG}
Jing Miao, Charat Thongprayoon, Supawadee Suppadungsuk, Oscar~A. Garcia~Valencia, and Wisit Cheungpasitporn.
\newblock Integrating retrieval-augmented generation with large language models in nephrology: Advancing practical applications.
\newblock {\em Medicina (Kaunas)}, 60(3):445, 2024.

\bibitem{Nakano2022}
Reiichiro Nakano, Jacob Hilton, Suchir Balaji, Jeff Wu, Long Ouyang, Christina Kim, Christopher Hesse, Shantanu Jain, Vineet Kosaraju, William Saunders, Xu~Jiang, Karl Cobbe, Tyna Eloundou, Gretchen Krueger, Kevin Button, Matthew Knight, Benjamin Chess, and John Schulman.
\newblock Webgpt: Browser-assisted question-answering with human feedback.
\newblock {\em arXiv preprint arXiv:2112.09332}, 2022.

\bibitem{Nogueira2019}
Rodrigo Nogueira and Kyunghyun Cho.
\newblock Passage re-ranking with bert.
\newblock In {\em Proceedings of the 2019 Conference of the North American Chapter of the Association for Computational Linguistics: Human Language Technologies (NAACL-HLT) (Demonstrations)}, pages 72--77, 2019.

\bibitem{Nori2023MedicalGPT}
Harsha Nori, Nicholas King, Scott Mayer~McKinney, Dean Carignan, and Eric Horvitz.
\newblock Capabilities of gpt-4 on medical challenge problems.
\newblock {\em arXiv preprint arXiv:2303.13375}, 2023.

\bibitem{nvidia2023rag}
NVIDIA.
\newblock Activate your data with custom generative ai.
\newblock {\em NVIDIA}, 2023.

\bibitem{Pan2023}
Yikang Pan, Liangming Pan, Wenhu Chen, Preslav Nakov, Min-Yen Kan, and William~Yang Wang.
\newblock On the risk of misinformation pollution with large language models.
\newblock In {\em Findings of the Association for Computational Linguistics: EMNLP 2023}, pages 1389--1403, 2023.

\bibitem{parvez-etal-2021-retrieval-augmented}
Md~Rizwan Parvez, Wasi Ahmad, Saikat Chakraborty, Baishakhi Ray, and Kai-Wei Chang.
\newblock Retrieval augmented code generation and summarization.
\newblock In {\em Findings of EMNLP 2021}, pages 2719--2734, 2021.

\bibitem{petroni2021kilt}
Fabio Petroni, Aleksandra Piktus, Angela Fan, Patrick Lewis, Majid Yazdani, Nicola De~Cao, James Thorne, Yacine Jernite, Vladimir Karpukhin, Jean Maillard, Vassilis Plachouras, Tim Rockt{\"a}schel, and Sebastian Riedel.
\newblock {KILT}: a benchmark for knowledge intensive language tasks.
\newblock In {\em Proceedings of the 2021 Conference of the North American Chapter of the Association for Computational Linguistics: Human Language Technologies}, pages 2523--2544, Online, June 2021. Association for Computational Linguistics.

\bibitem{petroni-etal-2021-kilt}
Fabio Petroni, Aleksandra Piktus, Angela Fan, Patrick Lewis, Majid Yazdani, Nicola De~Cao, James Thorne, Yacine Jernite, Vladimir Karpukhin, Jean Maillard, Vassilis Plachouras, Tim Rockt{\"a}schel, and Sebastian Riedel.
\newblock Kilt: a benchmark for knowledge intensive language tasks.
\newblock In {\em Proceedings of NAACL-HLT 2021}, pages 2523--2544, 2021.

\bibitem{pradeep2024ragnarok}
Ronak Pradeep, Nandan Thakur, Sahel Sharifymoghaddam, Eric Zhang, Ryan Nguyen, Daniel Campos, Nick Craswell, and Jimmy Lin.
\newblock Ragnar{\"o}k: A reusable {RAG} framework and baselines for the {TREC} 2024 retrieval-augmented generation track.
\newblock {\em arXiv preprint arXiv:2406.16828}, 2024.

\bibitem{Pradeep2024}
Ronak Pradeep, Nandan Thakur, Sahel Sharifymoghaddam, Eric Zhang, Ryan Nguyen, Daniel Campos, Nick Craswell, and Jimmy Lin.
\newblock {Ragnar\"ok}: A reusable retrieval‑augmented generation framework and baselines for the {TREC} 2024 {RAG} track.
\newblock {\em arXiv preprint arXiv:2406.16828}, 2024.
\newblock Introduces the AutoNuggetizer evaluation and shows its high correlation with human judgments.

\bibitem{radeva2024web}
Irina Radeva, Ivan Popchev, Lyubka Doukovska, and Miroslava Dimitrova.
\newblock Web application for retrieval-augmented generation: Implementation and testing.
\newblock {\em Electronics}, 13(7):1361, 2024.

\bibitem{Raffel2020}
Colin Raffel, Noam Shazeer, Adam Roberts, Katherine Lee, Sharan Narang, Michael Matena, Yanqi Zhou, Wei Li, and Peter~J. Liu.
\newblock Exploring the limits of transfer learning with a unified text-to-text transformer.
\newblock {\em Journal of Machine Learning Research}, 21(140):1--67, 2020.
\newblock Introduces the T5 model family, including T5-Large used as a closed-book QA baseline.

\bibitem{aws2023rag}
AWS~AI Research.
\newblock Enterprise ai with retrieval-augmented generation.
\newblock {\em AWS Blog}, 2023.

\bibitem{roberts2020}
Adam Roberts, Colin Raffel, and Noam Shazeer.
\newblock How much knowledge can you pack into the parameters of a language model?
\newblock In {\em EMNLP}, 2020.

\bibitem{sachan-etal-2021-end}
Devendra~Singh Sachan, Siva Reddy, William~L. Hamilton, Chris Dyer, and Dani Yogatama.
\newblock End-to-end training of multi-document reader and retriever for open-domain question answering.
\newblock {\em CoRR}, abs/2106.05346, 2021.

\bibitem{sachan2021emdr2}
Devendra~Singh Sachan, Siva Reddy, William~L. Hamilton, Chris Dyer, and Dani Yogatama.
\newblock End-to-end training of multi-document reader and retriever for open-domain question answering.
\newblock In {\em Proceedings of the 59th Annual Meeting of the Association for Computational Linguistics (ACL 2021)}, pages 2074--2088, Online, 2021. Association for Computational Linguistics.

\bibitem{salemi2024evaluating}
Alireza Salemi and Hamed Zamani.
\newblock Evaluating retrieval quality in retrieval-augmented generation.
\newblock In {\em Proceedings of the 47th International ACM SIGIR Conference on Research and Development in Information Retrieval}, pages 2395--2400, 2024.

\bibitem{Sankaradas2025}
Murugan Sankaradas, Ravi~K. Rajendran, and Srimat Chakradhar.
\newblock Streamingrag: Real-time contextual retrieval and generation framework.
\newblock {\em arXiv preprint arXiv:2501.14101}, 2025.

\bibitem{shuster2022bb3}
Kurt Shuster, Da~Ju, Peng Xu, Eric~Michael Smith, Emily Dinan, Stephen Roller, Piali Koura, Y-Lan Boureau, and Jason Weston.
\newblock Blenderbot 3: A deployed conversational agent that continually learns to responsibly engage.
\newblock {\em arXiv preprint arXiv:2208.03188}, 2022.

\bibitem{Shuster2022}
Kurt Shuster, Mojtaba Komeili, Leonard Adolphs, Stephen Roller, Arthur Szlam, and Jason Weston.
\newblock Language models that seek for knowledge: Modular search \& generation for dialogue and prompt completion.
\newblock In {\em Findings of the Association for Computational Linguistics: EMNLP 2022}, pages 373--393. Association for Computational Linguistics, 2022.

\bibitem{Shuster2021}
Kurt Shuster, Spencer Poff, Moya Chen, Douwe Kiela, and Jason Weston.
\newblock Retrieval augmentation reduces hallucination in conversation.
\newblock In {\em Findings of the Association for Computational Linguistics: EMNLP 2021}, pages 3784--3803, 2021.

\bibitem{singh2025agentic}
Aditi Singh, Abul Ehtesham, Saket Kumar, and Tala Talaei~Khoei.
\newblock Agentic retrieval-augmented generation: A survey on agentic rag.
\newblock {\em arXiv preprint arXiv:2501.09136}, 2025.

\bibitem{Singh2025ChunkRAG}
Ishneet~S. Singh, Ritvik Aggarwal, Ibrahim Allahverdiyev, Muhammad Taha, Aslihan Akalin, Kevin Zhu, and Sean O’Brien.
\newblock Chunkrag: A novel llm-chunk filtering method for rag systems.
\newblock {\em arXiv preprint arXiv:2410.19572}, 2025.

\bibitem{siriwardhana2023improving}
Shamane Siriwardhana, Rivindu Weerasekera, Elliott Wen, Tharindu Kaluarachchi, Rajib Rana, and Suranga Nanayakkara.
\newblock Improving the domain adaptation of retrieval augmented generation (rag) models for open domain question answering.
\newblock {\em Transactions of the Association for Computational Linguistics}, 11:1--17, 2023.

\bibitem{soudani2024fine}
Heydar Soudani, Evangelos Kanoulas, and Faegheh Hasibi.
\newblock Fine tuning vs. retrieval augmented generation for less popular knowledge.
\newblock In {\em Proceedings of the 2024 Annual International ACM SIGIR Conference on Research and Development in Information Retrieval in the Asia Pacific Region}, pages 12--22, 2024.

\bibitem{Sukhbaatar2015}
Sainbayar Sukhbaatar, Arthur Szlam, Jason Weston, and Rob Fergus.
\newblock End-to-end memory networks.
\newblock In {\em Advances in Neural Information Processing Systems 28 (NIPS 2015)}, pages 2440--2448, 2015.

\bibitem{Tang2024}
Yixuan Tang and Yi~Yang.
\newblock Multihop-rag: Benchmarking retrieval-augmented generation for multi-hop queries.
\newblock In {\em COLM}, 2024.
\newblock OpenReview preprint.

\bibitem{Harvey2025}
Harvey Team.
\newblock Enterprise-grade rag systems: High-performance rag with vector databases, 2025.
\newblock Harvey AI blog.

\bibitem{Trivedi2023}
Harsh Trivedi, Niranjan Balasubramanian, Tushar Khot, and Ashish Sabharwal.
\newblock Interleaving retrieval with chain-of-thought reasoning for knowledge-intensive multi-step questions.
\newblock In {\em Proceedings of the 61st Annual Meeting of the Association for Computational Linguistics (ACL)}, pages 10014--10037, 2023.

\bibitem{Veturi2024}
Sriram Veturi, Saurabh Vaichal, Reshma~Lal Jagadheesh, Nafis~Irtiza Tripto, and Nian Yan.
\newblock Rag based question-answering for contextual response prediction system.
\newblock In {\em Proceedings of the 33rd ACM International Conference on Information and Knowledge Management}, 2024.

\bibitem{wang2018r3}
Shuohang Wang, Mo~Yu, Xinya Guo, Zhiguo Wang, Tim Klinger, Wei Zhang, and Jing Jiang.
\newblock R\^{}3: Reinforced reader-ranker for open-domain question answering.
\newblock In {\em Proceedings of the 32nd {AAAI} Conference on Artificial Intelligence (AAAI)}, pages 5981--5988, 2018.

\bibitem{wang2024searching}
Xiaohua Wang, Zhenghua Wang, Xuan Gao, Feiran Zhang, Yixin Wu, Zhibo Xu, Tianyuan Shi, Zhengyuan Wang, Shizheng Li, Qi~Qian, et~al.
\newblock Searching for best practices in retrieval-augmented generation.
\newblock In {\em Proceedings of the 2024 Conference on Empirical Methods in Natural Language Processing}, pages 17716--17736, 2024.

\bibitem{wang2023learning}
Zhiruo Wang, Jun Araki, Zhengbao Jiang, Md~Rizwan Parvez, and Graham Neubig.
\newblock Learning to filter context for retrieval-augmented generation.
\newblock {\em arXiv preprint arXiv:2311.08377}, 2023.

\bibitem{Weston2015}
Jason Weston, Sumit Chopra, and Antoine Bordes.
\newblock Memory networks.
\newblock In {\em 3rd International Conference on Learning Representations (ICLR), Conference Track Proceedings}, 2015.

\bibitem{wu2025srrag}
Di~Wu, Jia-Chen Gu, Kai-Wei Chang, and Nanyun Peng.
\newblock Self-routing rag: Binding selective retrieval with knowledge verbalization.
\newblock {\em arXiv preprint arXiv:2504.01018}, 2025.

\bibitem{xiong2024benchmarking}
Guangzhi Xiong, Qiao Jin, Zhiyong Lu, and Aidong Zhang.
\newblock Benchmarking retrieval-augmented generation for medicine.
\newblock In {\em Findings of the Association for Computational Linguistics ACL 2024}, pages 6233--6251, 2024.

\bibitem{yang2025simrag}
Diji Yang, Linda Zeng, Jinmeng Rao, and Yi~Zhang.
\newblock Knowing you don't know: Learning when to continue search in multi-round rag through self-practicing.
\newblock In {\em Proceedings of SIGIR 2025}, 2025.
\newblock arXiv:2505.02811.

\bibitem{Yang2025RAGHealthcare}
Ruiyang Yang, Xin Huang, Xingyu Li, et~al.
\newblock Retrieval-augmented generation for generative artificial intelligence in health care.
\newblock {\em npj Health Systems}, 2(2):1--8, 2025.

\bibitem{yu2024rankrag}
Yue Yu, Wei Ping, Zihan Liu, Boxin Wang, Jiaxuan You, Chao Zhang, Mohammad Shoeybi, and Bryan Catanzaro.
\newblock Rankrag: Unifying context ranking with retrieval-augmented generation in llms.
\newblock {\em Advances in Neural Information Processing Systems}, 37:121156--121184, 2024.

\bibitem{Zeng2024}
Shenglai Zeng, Jiankun Zhang, Pengfei He, Yiding Liu, Yue Xing, Han Xu, Jie Ren, Yi~Chang, Shuaiqiang Wang, Dawei Yin, and Jiliang Tang.
\newblock The good and the bad: Exploring privacy issues in retrieval-augmented generation ({RAG}).
\newblock In {\em Findings of the Association for Computational Linguistics: ACL 2024}, pages 4505--4524. Association for Computational Linguistics, 2024.

\bibitem{zhai2024samrag}
Wenjia Zhai.
\newblock Self-adaptive multimodal retrieval-augmented generation.
\newblock {\em arXiv preprint arXiv:2410.11321}, 2024.

\bibitem{zhang2024interactive}
Ruichen Zhang, Hongyang Du, Yinqiu Liu, Dusit Niyato, Jiawen Kang, Sumei Sun, Xuemin Shen, and H~Vincent Poor.
\newblock Interactive ai with retrieval-augmented generation for next generation networking.
\newblock {\em IEEE Network}, 2024.

\bibitem{zhang2024raft}
Tianjun Zhang, Shishir~G Patil, Naman Jain, Sheng Shen, Matei Zaharia, Ion Stoica, and Joseph~E Gonzalez.
\newblock Raft: Adapting language model to domain specific rag.
\newblock In {\em First Conference on Language Modeling}, 2024.

\bibitem{zhao2024frag}
Danyang Zhao.
\newblock {FRAG}: Toward federated vector database management for collaborative and secure retrieval-augmented generation.
\newblock arXiv preprint arXiv:2410.13272, 2024.
\newblock \url{https://doi.org/10.48550/arXiv.2410.13272}.

\bibitem{zhao2024retrieval}
Penghao Zhao, Hailin Zhang, Qinhan Yu, Zhengren Wang, Yunteng Geng, Fangcheng Fu, Ling Yang, Wentao Zhang, Jie Jiang, and Bin Cui.
\newblock Retrieval-augmented generation for ai-generated content: A survey.
\newblock {\em arXiv preprint arXiv:2402.19473}, 2024.

\bibitem{Zhou2025}
Pengcheng Zhou, Yinglun Feng, and Zhongliang Yang.
\newblock Privacy-aware rag: Secure and isolated knowledge retrieval.
\newblock {\em arXiv preprint arXiv:2503.15548}, 2025.

\end{thebibliography}

\fussy 
\end{multicols}

\end{document}